\documentclass[sigconf,nonacm]{acmart}
\AtBeginDocument{%
  }

\setcopyright{none}
\copyrightyear{2026}
\acmYear{2026}
\acmDOI{XXXXXXX.XXXXXXX}
\acmConference[KDD '26]{Proceedings of the ACM SIGKDD Conference on Knowledge Discovery and Data Mining}{2026}{TBD}
\acmISBN{978-1-4503-XXXX-X/2026}

\usepackage[utf8]{inputenc} 
\usepackage[T1]{fontenc}    
\usepackage{hyperref}       
\usepackage{url}            
\usepackage{booktabs}       
\usepackage{amsfonts}       
\usepackage{nicefrac}       
\usepackage{microtype}      
\usepackage{xcolor}         
\usepackage{tcolorbox}
\usepackage{listings}
\usepackage{tcolorbox}
\usepackage{enumitem}
\usepackage{xcolor}
\usepackage{courier}
\usepackage{multirow}
\usepackage[table]{xcolor}
\usepackage{enumitem} 
\usepackage[belowskip=0pt, aboveskip=4pt]{caption}
\usepackage{wrapfig}
\usepackage{listings}
\usepackage{xcolor}
\usepackage{enumitem}
\tcbuselibrary{breakable}
\usepackage{booktabs}
\usepackage{tabularx} 
\usepackage{listings}
\usepackage{amsmath}
\usepackage{subcaption} 
\usepackage{wrapfig}
\usepackage{booktabs}
\usepackage{listings}
\lstdefinelanguage{json}{
    basicstyle=\ttfamily\small,
    numbers=left,
    numberstyle=\small,
    stepnumber=1,
    numbersep=8pt,
    showstringspaces=false,
    breaklines=true,
    frame=single,
    backgroundcolor=\color{gray!10},
    string=[s]{"}{"},
    stringstyle=\color{blue},
    comment=[l]{//},
    commentstyle=\color{green!50!black},
    morekeywords={true,false,null},
    keywordstyle=\color{red}
}

\newcommand{\ERData}{ERUnderstand}

\begin{document}

\title{ERUnderstand: Evaluating Vision-Language Models on Structured ER Diagrams}

%

\author{Ali Ansari}
\affiliation{%
  \institution{Temple University}
  \department{Department of Computer and Information Sciences}
  \city{Philadelphia}
  \state{PA}
  \country{USA}
  \postcode{19122}
}
\email{ali.ansari@temple.edu}

\author{Yasmin Mohammadi}
\affiliation{%
  \institution{Temple University}
  \department{Department of Computer and Information Sciences}
  \city{Philadelphia}
  \state{PA}
  \country{USA}
  \postcode{19122}
}
\email{yasmin.mohammadi@temple.edu}

\author{Farnoush Nili}
\affiliation{%
  \institution{Temple University}
  \department{Department of Computer and Information Sciences}
  \city{Philadelphia}
  \state{PA}
  \country{USA}
  \postcode{19122}
}
\email{farnoush.nili@temple.edu}

\author{Parsa Esmaeilkhani}
\affiliation{%
  \institution{Temple University}
  \department{Department of Computer and Information Sciences}
  \city{Philadelphia}
  \state{PA}
  \country{USA}
  \postcode{19122}
}
\email{parsa.esmaeilkhani@temple.edu}

\author{Longin Jan Latecki}
\affiliation{%
  \institution{Temple University}
  \department{Department of Computer and Information Sciences}
  \city{Philadelphia}
  \state{PA}
  \country{USA}
  \postcode{19122}
}
\email{latecki@temple.edu}

\author{Eduard Dragut}
\affiliation{%
  \institution{Temple University}
  \department{Department of Computer and Information Sciences}
  \city{Philadelphia}
  \state{PA}
  \country{USA}
  \postcode{19122}
}
\email{edragut@temple.edu}



\begin{abstract}
Entity-Relationship Diagrams (ERDs) are central to conceptual database design, yet they are typically available only as rendered images rather than machine-readable schemas, creating a bottleneck for AI-assisted database engineering tasks such as schema understanding, migration, documentation, and intelligent design assistance. We introduce \emph{\ERData}, the first large-scale benchmark for structured understanding of ER diagrams, comprising \textbf{2,960} diagrams collected from curated educational sources, real-world schemas, and synthetically generated examples spanning diverse domains, multiple ER notations (including Chen and Silberschatz-style), varying complexity levels, and Extended Entity--Relationship (EER) constructs. Each diagram is paired with a standardized JSON representation that enables fine-grained evaluation of entities, relationships, attributes, keys, weak entities, inheritance hierarchies, and other structural components. Evaluating state-of-the-art Vision--Language Models (VLMs), we find that while common ERD elements are recovered reliably (F1 $>$ 0.74), performance drops sharply on weak entities (as low as 0.28 F1), multivalued attributes (0.14 F1), and $n$-ary relationships (0.07 F1). Reasoning-augmented models improve overall Macro-F1 by 15--25\%, but remain sensitive to linguistic priors, spatial layout biases, and increasing diagram complexity. \ERData\ fills a critical gap in multimodal benchmarks by providing a standardized, executable-schema-focused resource for advancing visual-structural reasoning in database engineering. The benchmark, dataset, evaluation toolkit, and generation code are publicly available at \url{https://github.com/salinaria/ERUnderstand}.
\end{abstract}




\maketitle

\section{Introduction}
The growing integration of Artificial Intelligence (AI) into database systems is transforming how structured data is designed, queried, documented, and maintained~\cite{Zhou:2021:DatabaseMeetsAI,Li:2024:LLMDataManagement}. Recent advances include text-to-SQL generation, schema retrieval, database copilots, automated documentation, and multimodal query processing~\cite{leis2025still,Deng2021TURL,JoT24}. Despite this progress, one fundamental capability remains largely unexplored: enabling AI systems to understand conceptual database schemas represented as Entity--Relationship Diagrams(ERDs).

\begin{figure*}[t]
  \centering
  \begin{tcolorbox}[colback=blue!5!white,
                    colframe=blue!50!black,
                    width=\textwidth,
                    boxrule=0.4pt,
                    arc=4pt,
                    boxsep=2pt,
                    left=2pt,
                    right=2pt,
                    top=4pt,
                    bottom=4pt,]
    \begin{minipage}[c]{0.33\textwidth}
      \centering
      \includegraphics[width=\linewidth]{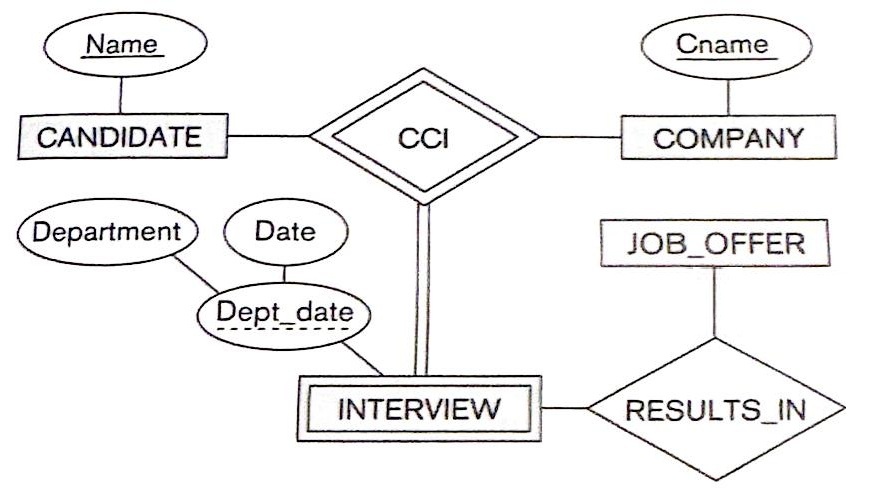}
    \end{minipage}%
    \hspace{0pt}
    \begin{minipage}[c]{0.34\textwidth}
      \centering
      \includegraphics[width=\linewidth]{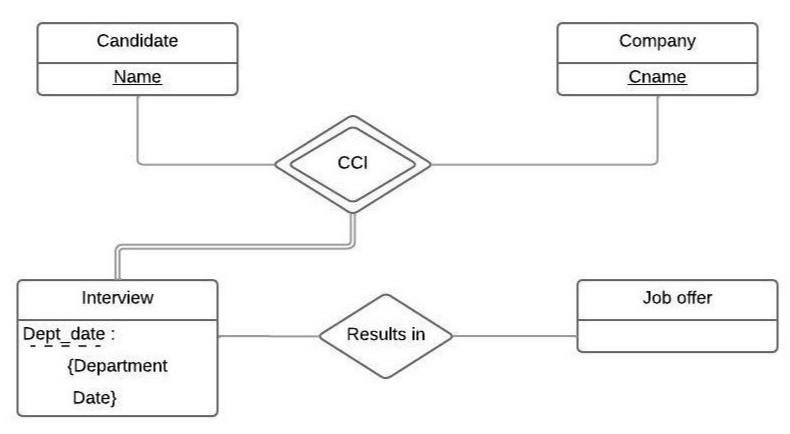}
    \end{minipage}%
    \hspace{0pt}
    \begin{minipage}[c]{0.30\textwidth}
      {\fontsize{7pt}{8pt}\selectfont
      
      \begin{itemize}[leftmargin=*, itemsep=2pt, topsep=0pt]
        \item \textcolor{blue!70!black}{\textbf{Ternary}} relationship: \texttt{``CCI''}.
        \item \textcolor{blue!70!black}{\textbf{Composite}} attribute: \texttt{``Dept\_date''}.
        \item \textcolor{blue!70!black}{\textbf{Weak entity}}: \texttt{``INTERVIEW''} (double line).
        \item \textcolor{blue!70!black}{\textbf{Weak relationship}}: \texttt{``CCI''} (double line).
        \item \textcolor{blue!70!black}{\textbf{Partial key}}: \texttt{``Dept\_date''} (dashed underline).
        \item \textcolor{blue!70!black}{\textbf{PKs}}: \texttt{``Name''}, \texttt{``Cname''} (underline).
      \end{itemize}
      }
    \end{minipage}
  \end{tcolorbox}
  \caption{A sample ERD shown in Chen's (left) and Silberschatz's notations (right).}
  \label{fig:elmasri}
  \label{fig:teaser}
  \vspace{-10pt}
\end{figure*}

ERDs remain the dominant abstraction for conceptual database design, capturing entities, relationships, integrity constraints, and inheritance structures prior to physical implementation~\cite{chen1976,thalheim2013}. In practice, however, these diagrams are rarely available in machine-readable form. Instead, they are predominantly distributed as rendered images embedded in textbooks, lecture slides, software documentation, PDFs, screenshots, collaborative platforms, and online repositories~\cite{ouyang2025omnidocbench,niu2025mineru2}. While database systems can deterministically process symbolic schema representations, they cannot directly interpret these visual artifacts. Recovering structured schemas from ERD images is therefore a fundamental capability for AI-assisted database engineering, enabling applications such as automated schema understanding, documentation generation, legacy system modernization, schema migration, and intelligent database assistants.

Recent Vision--Language Models (VLMs) have demonstrated remarkable capabilities in document understanding and multimodal reasoning, making them a natural candidate for extracting structured information from ER diagrams~\cite{gpt-report,gemini-report}. However, ERDs present challenges that differ substantially from those posed by natural images or conventional documents. Their semantics arise not only from textual labels, but also from formal graphical syntax, symbolic notation, spatial organization, and the topology of visual connections. Correct interpretation therefore requires jointly recovering diagram elements, their semantic roles, and the structural relationships that define the underlying conceptual schema, rather than simply recognizing objects or reading text.

Despite rapid progress in multimodal reasoning, existing benchmarks focus primarily on charts, flowcharts, scientific figures, geometric diagrams, and document layouts~\cite{pan2024flowlearn,lu2021iconqa,Engelhardt}. To the best of our knowledge, no benchmark systematically evaluates the structured understanding of conceptual database schemas represented as ER diagrams. Existing datasets emphasize visual perception or high-level document comprehension, whereas database engineering applications require reconstructing executable, machine-readable schema representations that can be directly consumed by downstream tools. This disconnect has limited our ability to measure and improve multimodal reasoning for conceptual database design.

To address this gap, we introduce \ERData, the first large-scale benchmark for structured understanding of Entity--Relationship Diagrams (ERDs). The benchmark comprises \textbf{2,960} diagrams collected from three complementary sources: (i) educational ERDs curated from authoritative textbooks and instructional materials, covering multiple ER notations and rich conceptual modeling constructs; (ii) production schemas automatically extracted from widely used database benchmarks and real-world systems, including Spider, BIRD, Northwind, AdventureWorks, Sakila, Chinook, Pagila, Employees, ClassicModels, and representative property-graph schemas; and (iii) synthetically generated ERDs produced through an LLM-assisted Graphviz pipeline, enabling controlled variation in domain, notation, structural complexity, and Extended Entity--Relationship (EER) constructs.

Each diagram is paired with a standardized machine-readable JSON representation, enabling fine-grained evaluation of schema recovery at the level of entities, relationships, attributes, keys, weak entities, inheritance hierarchies, and other structural components. Unlike answer-based evaluation, this representation allows precise analysis of where multimodal systems succeed or fail when reconstructing conceptual database schemas.

ERDs present a uniquely demanding multimodal reasoning task. Correct interpretation requires substantially more than recognizing graphical symbols or reading text: models must jointly infer topology, semantic roles, notation-dependent conventions, and structural constraints while preserving the integrity of the underlying conceptual graph. Figure~\ref{fig:elmasri} illustrates this challenge. Although the diagram contains only a handful of entities, accurately recovering the schema requires reasoning over graphical syntax, spatial connectivity, symbolic notation, and relational structure. As our experiments demonstrate, even state-of-the-art Vision--Language Models frequently recognize individual components while failing to reconstruct the complete conceptual schema.

Unlike existing multimodal benchmarks that primarily evaluate visual question answering, document understanding, or caption generation, \ERData\ evaluates the recovery of executable symbolic database schemas, making it directly relevant to AI-assisted database engineering. Specifically, this paper makes the following contributions:

\begin{itemize}[topsep=2pt,itemsep=1pt,leftmargin=*]

\item \textbf{The \ERData~ Benchmark.} We introduce the first large-scale benchmark for structured understanding of ER diagrams, comprising 2,960 diagrams spanning educational examples, production database schemas, and synthetically generated ERDs across diverse domains, multiple ER notations, and advanced EER constructs.

\item \textbf{A Standardized Evaluation Framework.} We provide a unified JSON schema, high-quality annotations, structure-aware evaluation metrics, an open-source evaluation toolkit, and reproducible benchmark protocols for fine-grained assessment of schema recovery.

\item \textbf{Comprehensive Benchmarking of VLMs.} We establish the first systematic benchmark of state-of-the-art Vision--Language Models on conceptual database diagram understanding, providing strong baselines for future research.

\item \textbf{Insights into Multimodal Structural Reasoning.} Through extensive analysis, we identify persistent failure modes in current VLMs—including higher-order relationships, notation-dependent semantics, spatial reasoning, and increasing structural complexity—highlighting key challenges for future multimodal systems operating on conceptual database schemas.

\end{itemize}

\begin{figure*}[t]
    \centering
    \includegraphics[width=.9\linewidth]{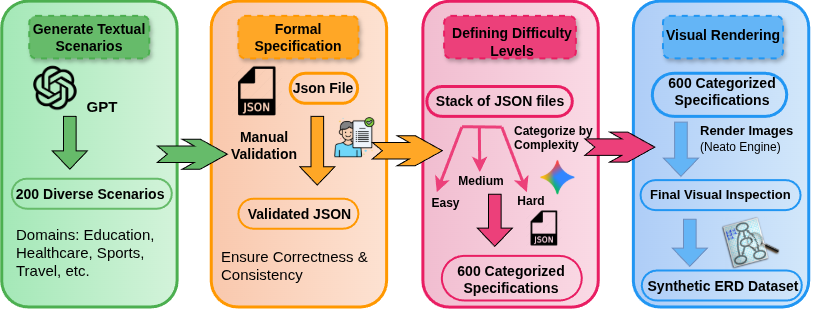}
    \caption{Four-stage pipeline for synthetic ERD generation.}
    \label{fig:pipeline}
    \vspace{-10pt}
\end{figure*}

\section{Related Work}
\label{sec:relatedwork}
Diagrams encompass diverse visual representations; we focus on \emph{diagrams as structured visual languages}, where graphical elements follow conventions that support logical reasoning \cite{barwise1995heterogeneous,jamnik2001,stenning1995}. Since Chen’s introduction of the ER model \cite{chen1976}, ERDs have served as formal languages with well-defined syntax and semantics \cite{giunchiglia1992theory,shimojima2015semantic}. Their precise encoding of database schemas makes them a compelling testbed for distinguishing  visual-structural reasoning from pattern matching in VLMs.

Recent work has evaluated VLMs on a range of diagram understanding tasks. IconQA \cite{lu2021iconqa} studies abstract visual reasoning, while CSDia \cite{wang2022computer} and flowchart benchmarks \cite{feng2023genflowchart,shen2024flowlearn,zhang2024first} focus on geometric and procedural diagrams. In parallel, SO-Bench, ExtractBench, OmniSch, and recent VLM-to-BPMN JSON extraction pipelines evaluate schema-grounded or diagram-to-graph structured output in UI, document, chart, PCB schematic, and workflow settings. OmniSch is especially close in spirit because it compares VLMs against a classical OCR/detector/heuristic baseline for schematic-to-netlist reconstruction. These efforts are complementary: they do not target ER/EER semantics such as weak entities, identifying relationships, $n$-ary relationships, and inheritance hierarchies, which require distinct annotation and evaluation machinery.

Within the database community, diagrammatic representations have long been central to schema design and query interfaces, including Query by Diagram \cite{angelaccio1990query}, visual query systems \cite{siau1998visual}, and tools such as ViziQuer \cite{ovcinnikiva2023viziquer}. Prior work has also explored visual encodings of SQL queries \cite{gatterbauer2011databases,leventidis2020queryvis}. However, despite this extensive literature, the ability of modern VLMs to interpret ERDs as structured data modeling artifacts has not been systematically studied.

\section{The \ERData\ Benchmark}
\label{dataset}

\ERData\ is a benchmark for evaluating the ability of VLMs to recover structured database schemas from ERD images. Unlike existing diagram and document understanding benchmarks that focus primarily on recognition or question answering, \ERData\ evaluates structured schema reconstruction by comparing model outputs against machine-readable ground-truth representations. Each ERD is paired with a standardized JSON schema encoding entities, attributes, relationships, keys, and EER constructs, enabling fine-grained evaluation of individual structural components under a unified scoring framework.

The benchmark contains \textbf{2,960 ERDs} collected from three complementary sources: (1) curated conceptual ERDs from educational materials and online repositories, (2) real database schemas extracted from widely used benchmarks and open-source systems, and (3) synthetic ERDs generated through an LLM-assisted Graphviz pipeline. Together, these sources provide complementary evaluation settings covering conceptual modeling, realistic database structures, and controlled stress-testing of multimodal structural reasoning. Table \ref{tab:stats} shows the statistics of the \ERData\ dataset. 

\subsection{Curated ERD Collection}

\textbf{Data Collection and Representation.}
We collected 182 ER diagrams from publicly available educational resources, including database textbooks, course materials, and online repositories. After removing duplicates and manually filtering low-quality examples, the collection contains 148 unique high-quality conceptual ERDs. To ensure reproducibility while respecting copyright restrictions, we release source references and metadata rather than redistributing the original images.

To evaluate robustness across visual representations, each underlying ERD schema is rendered using multiple graphical styles, including original source representations and standardized Graphviz renderings. This results in 740 diagram instances from the curated collection, covering multiple ER notations such as Chen and Silberschatz-style variants.

\textbf{Structured Annotation.}
Each diagram is converted into a unified JSON representation describing entities, attributes, relationships, cardinalities, primary keys, weak entities, identifying relationships, inheritance hierarchies, and other EER constructs. This representation enables automatic regeneration of diagrams and supports element-level evaluation beyond surface-level text matching.

\textbf{Annotation Quality.}
To measure annotation reliability, we conducted a double-annotation study on 30 curated ERDs (15 web-sourced and 15 textbook diagrams). Two annotators independently converted each diagram into the benchmark JSON format using the same annotation guidelines. Agreement was measured using the structure-aware Macro-F1 metric used throughout the benchmark evaluation, resulting in an average agreement score of 0.897; exactly 21 diagrams required joint review, primarily due to ambiguous notation or low-resolution images. Only finalized annotations are included in the released benchmark.

\begin{table*}[t]
\centering
\caption{Statistics of the \ERData\ benchmark. Educational ERDs include five visual representations for each underlying schema. Production schemas are rendered in a single standardized Graphviz notation.}
\label{tab:stats}
\setlength{\tabcolsep}{1.6pt}
\small
\begin{tabular}{lcccccccccccc}
\toprule
\textbf{Category} & \textbf{Num} & \textbf{Ents} & \textbf{Rels} & \textbf{Attrs} & \textbf{W.Ent} & \textbf{W.Rel} & \textbf{N-ary} & \textbf{IS-A} & \textbf{Deriv} & \textbf{Comp} & \textbf{Mult} & \textbf{PKs} \\
\midrule
\rowcolor{gray!10}Web & 500 & 2,955 & 2,625 & 10,640 & 80 & 45 & 150 & 85 & 45 & 220 & 125 & 1,320 \\
\rowcolor{gray!10}Instructional & 240 & 1,235 & 820 & 2,785 & 100 & 110 & 70 & 210 & 5 & 80 & 55 & 670 \\
\midrule
\rowcolor{yellow!15}Production Schemas & 179 & 1,036 & 995 & 5,426 & 0 & 0 & 0 & 0 & 0 & 0 & 0 & 836 \\
\midrule
\rowcolor{cyan!10}Easy & 597 & 2,409 & 2,325 & 8,445 & 441 & 447 & 432 & 0 & 0 & 0 & 0 & 2,490 \\
\rowcolor{cyan!10}Medium & 796 & 4,048 & 3,776 & 18,536 & 920 & 860 & 656 & 36 & 644 & 744 & 748 & 4,188 \\
\rowcolor{cyan!10}Hard & 597 & 4,533 & 4,377 & 19,530 & 1,164 & 927 & 546 & 156 & 522 & 612 & 579 & 4,617 \\
\rowcolor{cyan!10}High IS-A & 51 & 1,199 & 387 & 1,494 & 80 & 62 & 29 & 283 & 16 & 25 & 21 & 392 \\
\midrule
\textbf{Total} & \textbf{2,960} & \textbf{17,415} & \textbf{15,305} & \textbf{66,856} & \textbf{2,785} & \textbf{2,451} & \textbf{1,883} & \textbf{770} & \textbf{1,232} & \textbf{1,681} & \textbf{1,528} & \textbf{14,513} \\
\bottomrule
\end{tabular}
\end{table*}

\subsection{Real Database Schemas}

To complement conceptual ERDs, \ERData\ includes 179 real database schemas automatically extracted from widely used database benchmarks and open-source systems. The collection includes Spider, BIRD, Northwind, AdventureWorks, Sakila, Chinook, Pagila, Employees, ClassicModels, and representative property-graph schemas from domains such as recommendation systems, social networks, fraud detection, and knowledge graphs.

Unlike conceptual ERDs, these schemas are derived from existing database structures and therefore primarily contain entities, attributes, primary keys, and binary relationships, with limited representation of advanced EER constructs. All schemas are converted into the unified JSON representation and rendered using a standardized Graphviz style. This tier provides realistic database structures with an average of 46.4 structural elements per diagram, complementing the conceptual modeling focus of the curated collection.

\subsection{Synthetic ERD Generation}

To provide controlled evaluation across different levels of structural complexity, we develop a scalable synthetic generation pipeline grounded in the distributions observed in the curated collection. The released benchmark uses fixed random seeds for reproducibility, while the generation framework enables future extensions and anti-overfitting evaluations.

The generation process consists of four stages: generating domain-specific database scenarios, converting scenarios into structured JSON schemas, rendering ER diagrams using Graphviz, and assigning difficulty levels based on structural complexity and EER characteristics. The synthetic collection systematically varies schema size, relationship arity, hierarchy depth, spatial layout, and advanced constructs including weak entities, identifying relationships, inheritance hierarchies, composite attributes, and multivalued attributes.

The synthetic subsets are designed to complement the limitations of the curated and production collections. Easy and medium subsets follow the scale and complexity distribution of common educational ERDs, while hard subsets introduce substantially larger schemas with richer EER structures. By increasing the frequency of challenging constructs that occur less frequently in real collections, the synthetic component enables targeted analysis of VLM capabilities under controlled conditions.

Unlike static benchmarks, the synthetic generation pipeline allows new schemas, domains, and structural variations to be produced while maintaining the same annotation and evaluation framework. This enables reproducible testing of multimodal reasoning capabilities across factors such as diagram complexity, topology, and notation variation. Figure~\ref{fig:pipeline} illustrates the generation pipeline.

\section{Evaluation Methodology}
\label{evalMethod}
We evaluate ERD understanding using two complementary metrics: a structure-aware fractional F1-score for fine-grained correctness and BLEU for sequence-level similarity, capturing both  structural accuracy and overall output quality.

\textbf{Fractional F1-Score.}
Our main metric is a fractional F1-score, designed for structured outputs with interdependent elements. Unlike standard F1~\cite{chinchor}, it accounts for \textit{partial correctness}, allowing partially correct predictions. Structural errors (e.g., incorrect relationships, cardinalities, or attribute assignments) are assigned 0.5 weight as both False Positive (FP) and False Negative (FN), following a standard partial-credit heuristic. We use greedy matching to align predicted elements with ground truth based on label similarity before evaluating structural properties; Appendix~\ref{sec:experimental_details} gives prompting, normalization, and a worked $n$-ary example. We compute F1 separately for entities, relationships, attributes, weak entities/relationships, and primary keys. Relationship scoring accounts for missed, extra, misconnected, and incorrect predictions:

\begin{align}
TP_{\text{rel}} &= \text{\#relations} - \text{Rel}_{\text{missed}} - \text{Rel}_{\text{miscon}} - \text{N-ary} \\
FP_{\text{rel}} &= \text{Rel}_{\text{extra}} + \frac{\text{Rel}_{\text{miscon}} + \text{Cardinality} + \text{N-ary}}{2} \\
FN_{\text{rel}} &= \text{Rel}_{\text{missed}} + \frac{\text{Rel}_{\text{miscon}}}{2}, \quad
\text{F1}_{\text{rel}} = \frac{2TP_{\text{rel}}}{2TP_{\text{rel}} + FP_{\text{rel}} + FN_{\text{rel}}}
\end{align}

Invalid or unparsable JSON outputs are scored as empty predictions; we additionally track valid-JSON coverage separately in Appendix~\ref{sec:experimental_details}.

\textbf{BLEU and Graph Edit Distance (GED).} We use BLEU~\cite{papineni2002bleu} to measure similarity between predicted and ground-truth JSON outputs, providing a sequence-level complement to structural evaluation. To reduce sensitivity to ordering, we deterministically sort entities, relationships, and attributes before computing BLEU with NLTK~\cite{bird2009natural}. We additionally evaluate predictions using Graph Edit Distance (GED)~\cite{GaoXTL10}, which measures the minimum number of graph edit operations required to transform a predicted graph into the ground-truth graph. While GED provides a topology-aware metric, we found that it offers limited additional insight beyond Fractional F1, as small local mismatches can propagate into disproportionately large penalties in dense ERDs. We therefore use GED as a complementary structural metric rather than a primary evaluation measure.

\begin{figure*}[t]
    \centering
    \includegraphics[width=0.5\linewidth]{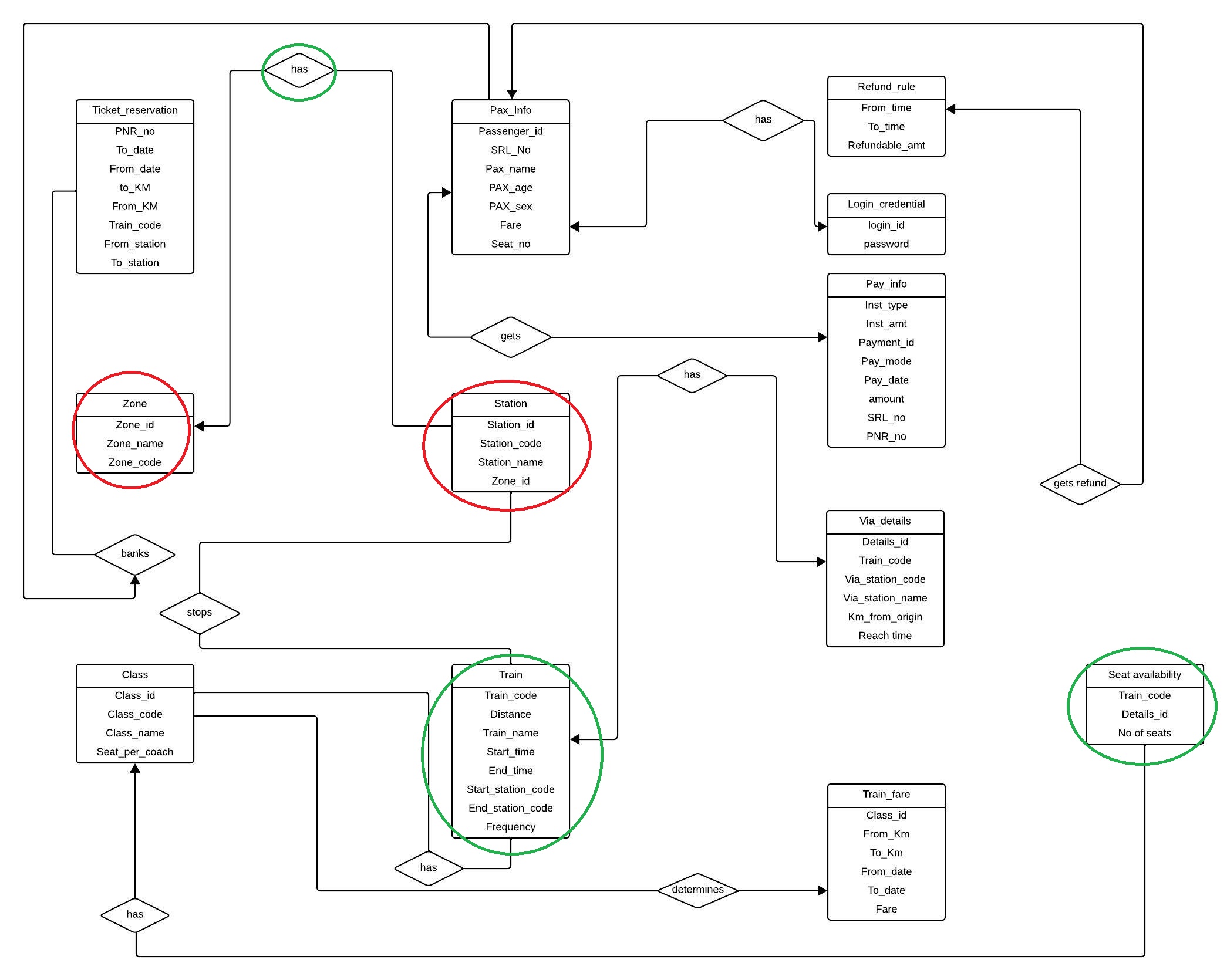}
    \hspace{0.05\linewidth}
    \includegraphics[width=0.35\linewidth]{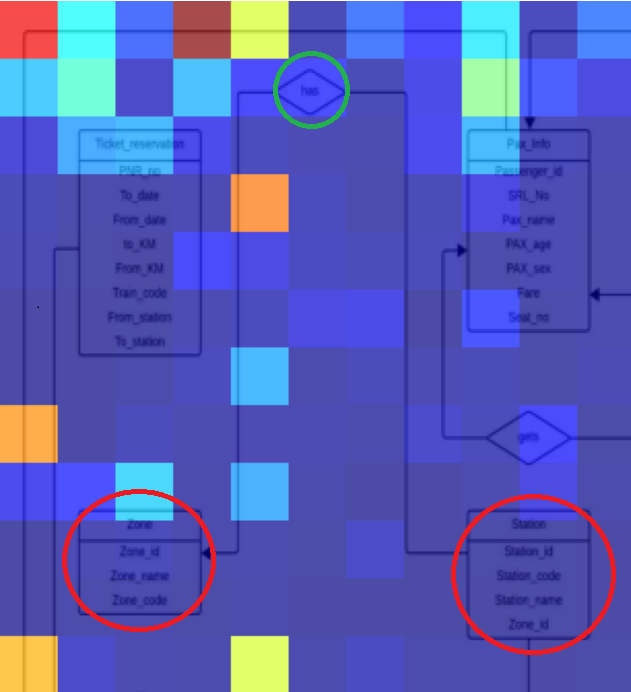}
    \caption{(a) An example test ER image. (b) The attention map of the top left part for the query
    \emph{What entities is 'has' connected to directly?}, generated in the way similar to \cite{Aflalo_2022_CVPR}.}
    \label{fig:attentionCrop}
\vspace{-5pt}
\end{figure*}

\section{Failure Mode Analysis}
\label{sec:error}
We analyze the failure modes underlying aggregate performance through controlled experiments. We identify three recurring structural biases (i.e., spatial bias, linguistic prior dependence, and complexity collapse) that appear consistently across both open- and closed-source models. These patterns suggest recurring weaknesses shared across model families rather than vendor-specific behavior and provide context for the quantitative results in Section \ref{experiments}.

\textbf{Spatial Bias: Insensitivity to Visual Fidelity.}
\label{sec:present}
We observe a consistent performance gap between curated and synthetic ERDs, with curated diagrams underperforming despite higher visual quality (Macro-F1: $0.59$ vs. $0.67$; BLEU: $0.35$–$0.41$ vs. $0.40$–$0.48$). To isolate the effect of visual fidelity, we regenerate curated ERDs from the same underlying JSON using Graphviz, producing semantically identical diagrams in three notational styles. 
If visual fidelity were the dominant factor, these renderings would improve performance. Instead, results remain largely unchanged across notations. 
For Web ERDs, Macro-F1 varies narrowly between $0.62$ and $0.66$, while Instructional ERDs range from $0.60$ to $0.67$. Component-level performance is similarly stable, with entity extraction reaching up to $0.97$, relationship detection up to $0.85$, and attribute recognition up to $0.94$, independent of notation.
These results indicate that visual fidelity does not explain the curated–synthetic gap, motivating analysis of semantic and structural factors.

\textbf{Linguistic Prior Dependence.}
To assess whether semantic context influences ERD understanding, we evaluate the synthetic dataset across 10 thematic categories spanning 20 domains, testing whether VLMs rely on domain familiarity to compensate for weaknesses in structural reasoning.
Across all categories, Macro-F1 exhibits minimal variance. In the Easy tier, scores range from $0.77$ (Legal) to $0.83$ (Education), averaging $0.80$, with similar stability in Medium and Hard settings ($0.72$ and $0.67$). No domain consistently outperforms others. These results indicate that ERD interpretation errors are largely independent of semantic domain, revealing limitation in structural reasoning than reliance on domain knowledge.

\textbf{Complexity Degradation.}
\label{sec:size}
To test scalability, we evaluate models on highly complex ERDs (20+ entities and relationships), using a subset of 19 diagrams meeting this criterion. Performance degrades across most models, with Macro-F1 dropping to $0.035$–$0.048$ and relationship scores near zero (as low as $0.009$). In contrast, Gemini models remain substantially more robust: Gemini-2.0 achieves $0.428$ Macro-F1, while Gemini-2.5 reaches $0.695$ with strong performance across entities and relationships. These results show a clear scalability limit in most VLMs, with sharp degradation beyond a complexity threshold, while Gemini models maintain significantly higher performance on dense visual structures.

\textbf{Spatial Proximity Bias in ERDs.} Observations 1 and 2 reveal a consistent gap between entity and relationship understanding: models achieve high F1-scores for entities (0.90) but perform substantially worse on relationships (15\% drop on curated data, 10\% on synthetic). Error analysis shows two recurring patterns: hallucinated relationships between spatially proximate entities (false positives) and missed relationships between distant but correctly connected entities (false negatives), indicating failures in spatial reasoning rather than semantics.

We hypothesize that models exhibit spatial proximity bias inferring relationships based on visual closeness while ignoring diagram logic and missing distant yet valid connections. Figure~\ref{fig:attentionCrop}(a) illustrates this effect: Qwen hallucinates the relation (Train has Seat Availability), with entities and relation name shown in green circles, while ignoring the true connections (Zone has Station).  The hallucinated relation is semantically plausible (``Train has seat availability''), whereas the correct relation is visually grounded, suggesting reliance on superficial cues.

\begin{figure}[t]
    \centering
    \includegraphics[width=\linewidth]{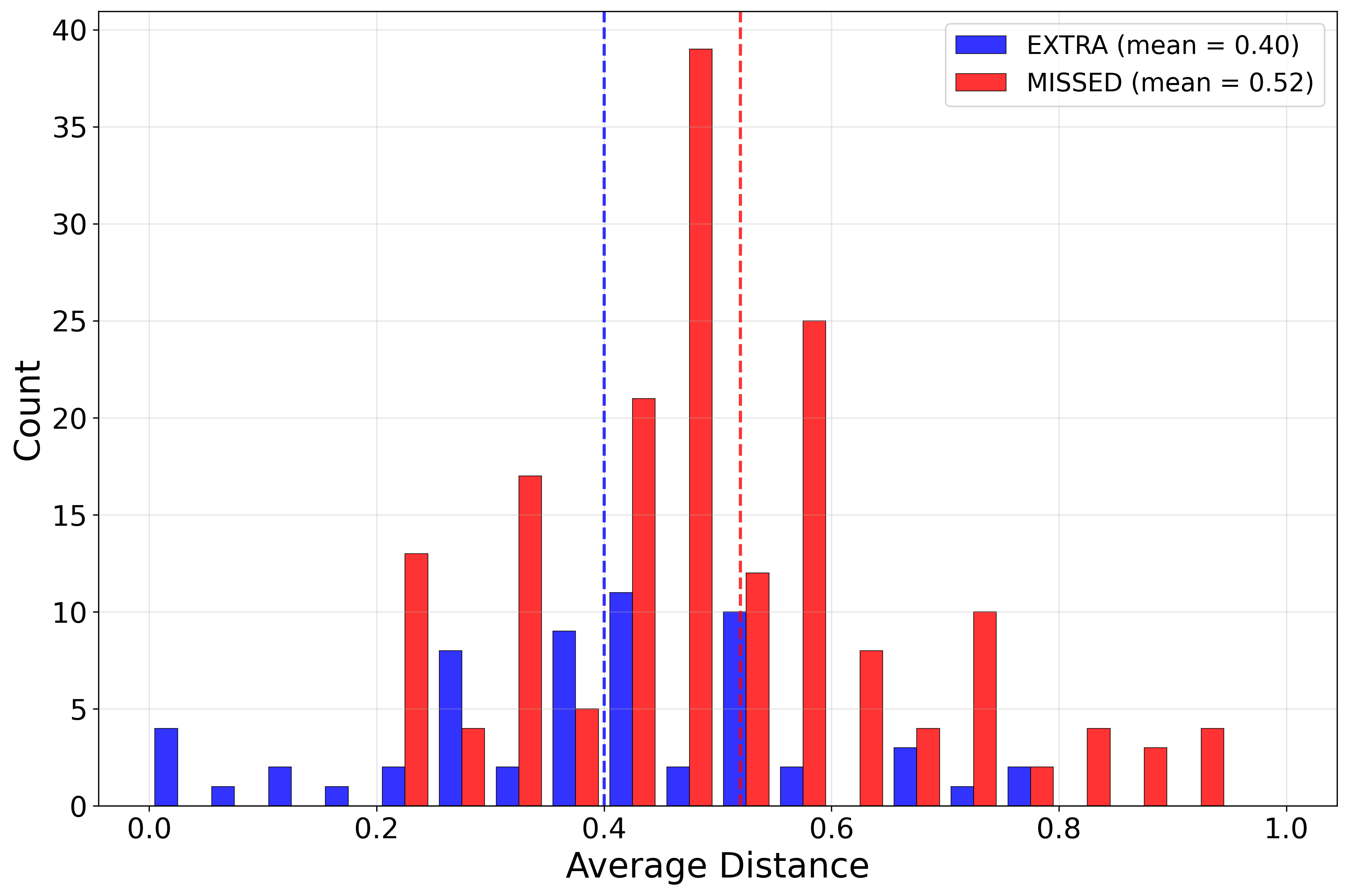}
    \caption{Distribution of distances for extra and missed relationship errors, highlighting spatial localization.}
    \label{fig:distdistr}
    \vspace{-20pt}
\end{figure}

To verify this hypothesis, we analyze the attention patterns using Qwen. Figure~\ref{fig:attentionCrop}(b) shows the attention map for the query \emph{What entities are directly connected to `has'?} The model fails to attend to the entity \textit{Station} or the connecting edge, suggesting limited attention to the relevant connecting edges and entities. This bias is further supported by distance analysis. As shown in Figure~\ref{fig:distdistr}, hallucinated relationships (EXTRA) occur predominantly between nearby entities, while missed relationships (MISSED) are more likely between distant ones. The separation in mean distances (dashed lines) highlights this divergence.

These findings suggest that current VLMs rely heavily on spatial heuristics rather than explicit visual connectivity. Addressing this limitation likely requires architectural improvements in modeling explicit visual connectivity rather than more training data.

\textbf{Prior-Language Bias in ERDs.}
\label{sec:language}
The spatial bias analysis shows that models hallucinate relationships between nearby entities while missing valid distant ones. However, distance alone cannot explain why some hallucinations are preferred. For example, the relation \textit{Train has Seat Availability} is inferred while \textit{Zone has Station} is ignored despite clear visual links (Figure~\ref{fig:attentionCrop}(a)), suggesting reliance on semantic plausibility.

We hypothesize that VLMs depend on prior-language information, like familiar entity names, relationship patterns, and co-occurrence statistics, when interpreting ERDs, reflecting reliance on learned visual–linguistic associations \cite{gpt-report,gemini-report} rather than fully grounded structural reasoning. To isolate this effect, we design naming strategies disrupting linguistic cues while preserving structure:

\begin{itemize}[topsep=0pt, partopsep=0pt, itemsep=0pt, parsep=0pt, leftmargin=*]
    \item \textbf{Context-Free (CF)}: Replace all labels with random random 4-letter strings (e.g., zebc, muxz, aydf). This strategy aims to remove semantic content, encouraging VLMs to rely solely on the semantic content conveyed through diagrammatic notation rather than linguistic cues.
       
    \item \textbf{Label Permutation (LP)}: Randomly shuffle labels across entities, attributes, and relationships, preserving vocabulary but breaking semantic alignment. For example, a relationship label such as \textit{triggers} may be reassigned as an entity name, while \textit{username} may become a relationship label. This manipulation disrupts label-role alignment and serves as a controlled probe of reliance on linguistic cues.
\end{itemize}

\begin{figure}[t]
    \centering
    \includegraphics[width=\linewidth]{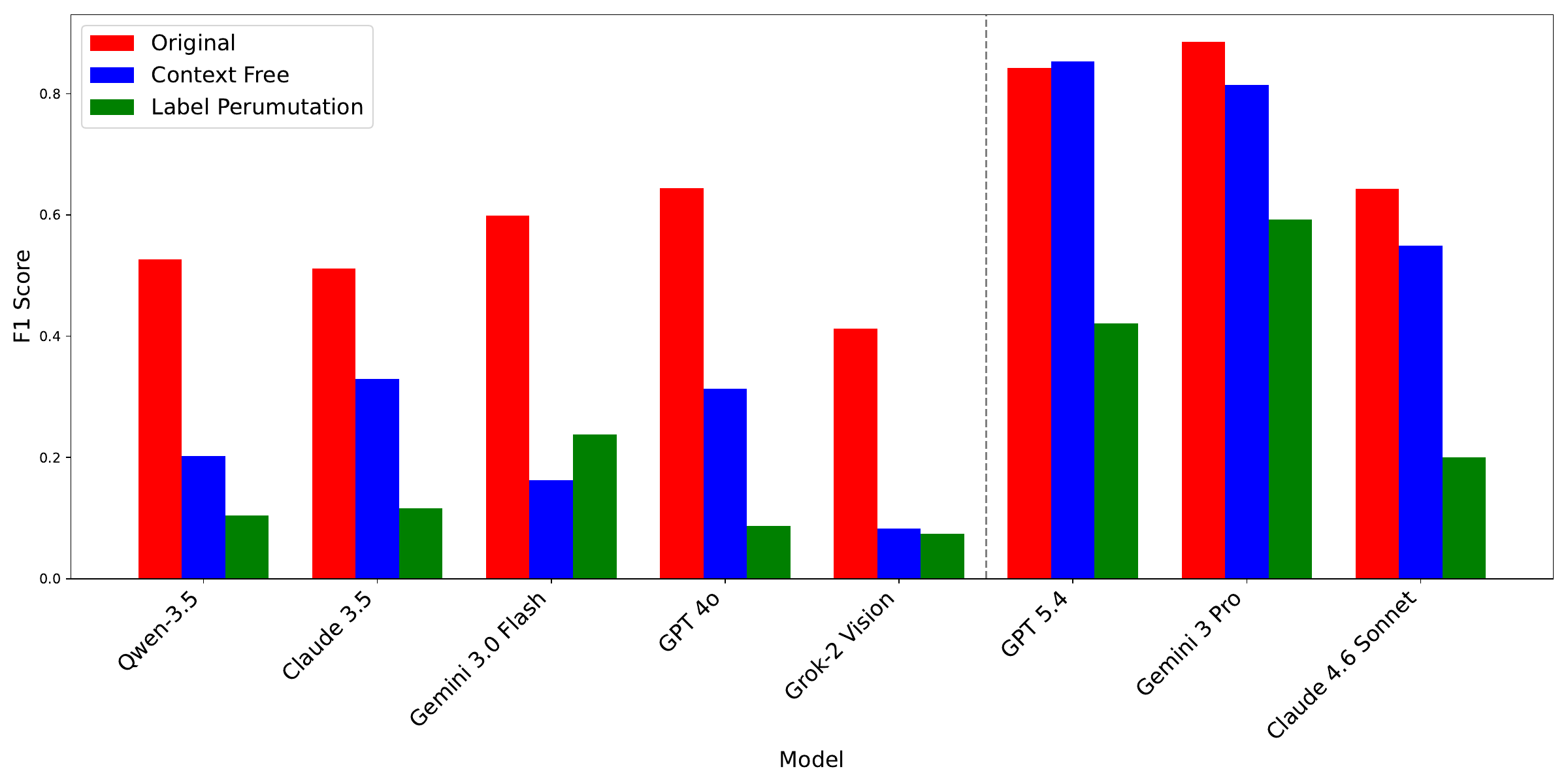}
    \caption{F1-scores across Original, CF, and LP settings. Reasoning models show smaller drops under label perturbations, suggesting greater robustness to disrupted contextual cues.}
    \label{fig:f1-biased}
    \vspace{-10pt}
\end{figure}

We evaluate both settings on 200 medium-complexity ERDs. Figure~\ref{fig:f1-biased} reports average F1 per diagram. Under CF (blue bars), all models exhibit significant performance drops, confirming strong reliance on linguistic cues. Reasoning models (right of the dashed line) are more robust, with smaller declines (e.g., GPT-5.4: $-0.08$, Claude-4.6: $-0.20$), while non-reasoning models degrade substantially (e.g., Claude-4.5: $-0.35$, Gemini-3.0-flash: $-0.65$). Under LP (red bars), performance again declines across all models, indicating sensitivity to disrupted label-role alignment. Drops range from $-0.27$ (Gemini-3.0) to $-0.49$ (Claude-4.6), with non-reasoning models showing similarly large degradation (e.g., Qwen-3.5 $-0.51$).

These results demonstrate a strong dependence on prior-language information for inferring ERD structure. While reasoning models are more resilient, most VLMs rely heavily on linguistic patterns rather than diagrammatic structure, limiting generalization to unfamiliar domains or naming conventions. Notably, GPT-5.4’s relatively stable performance across conditions suggests greater robustness to prior-language perturbations and a stronger reliance on diagrammatic structure.

\textbf{Hierarchy Tracing Failure.}
\label{sec:IS-A}
We observe a sharp drop in IS-A relationship performance when semantic cues are removed. To isolate visual hierarchy tracing, we construct IS-A-only diagrams containing only superclass--subclass links. This subset includes 32 diagrams. 
Performance degrades sharply across most models: for example, GPT-4o drops from 0.96 to 0.12 and Gemini-2.5 from 0.94 to 0.07, while entity recognition remains high. This indicates that models can detect entities but struggle to infer correct hierarchy direction. Overall, results suggest that IS-A predictions in standard ERDs rely heavily on contextual and semantic cues, degrading significantly when models must rely purely on visual structure.

\textbf{Primary Key Bias: Naming vs.\ Visual Cues.}
\label{sec:primary}
We investigate whether models identify primary keys (PKs) using visual cues (e.g., underlines) or naming conventions (e.g., \texttt{\_id}). Controlled experiments show strong dependence on naming patterns. Altering naming conventions substantially degrades performance (missed PKs increase from 40.75 to 105.50), while removing visual cues leads to widespread hallucination (up to 479.62 extra PKs). Misleading naming further amplifies errors (979.88 missed, 799.25 extra). These results indicate a bias toward naming conventions over visual notation in PK identification. This explains the gap between synthetic and curated datasets: models fail to reliably interpret visual PK cues, reducing robustness in real-world ERDs.

\textbf{Limits of Interactive Prompting.}
To assess whether VLM errors can be mitigated through interaction, we evaluate iterative prompting. We select 30 diagrams spanning multiple difficulty levels using three models (Grok, Gemini, and Claude), We use a standardized initial prompt followed by targeted follow-up queries. To avoid biasing model responses, we adopt an indirect questioning strategy (e.g., asking models to list weak relationships instead of confirming specific ones). This encourages independent reassessment rather than reliance on suggested corrections. Interactive prompting improves performance across all models, particularly for missed elements. For example, Grok reduces missed attributes by 25.9\% and missed relationships by 57.1\%, while Claude achieves the largest overall error reduction (39.9\%), including an 84.2\% decrease in hallucinated primary keys. However, improvements are not uniform. Errors related to structural reasoning, such as incorrect connections and misinterpretation of visual notation (e.g., multivalued attributes), persist after interaction, and in some cases, additional dialogue introduces new errors. Overall, interactive prompting helps recover overlooked elements but has limited impact on deeper reasoning failures, suggesting that while dialogue refines attention, it cannot compensate for limited spatial and structural understanding.

\begin{table*}[t]
    \centering
    \caption{BLEU and Macro-F1 scores on synthetic and curated ERD datasets. Web-S and Instr.-S denote Silberschatz-style renderings. Gray: non-reasoning, Cyan: reasoning models. \textsuperscript{\dag} indicates open-source. Best scores in bold.}
    \label{tab:final_scores}
    
    \setlength{\tabcolsep}{2.2pt} 
    \renewcommand{\arraystretch}{1.1}
    
    \begin{tabular}{lccccccccccccccccc}
        \toprule
        \textbf{Model} & \multicolumn{2}{c}{\textbf{Easy}} & \multicolumn{2}{c}{\textbf{Med}} & \multicolumn{2}{c}{\textbf{Hard}} & \multicolumn{2}{c}{\textbf{Web}} & \multicolumn{2}{c}{\textbf{Web-S}} & \multicolumn{2}{c}{\textbf{Instr.}} & \multicolumn{2}{c}{\textbf{Instr.-S}} & \multicolumn{2}{c}{\textbf{Schema}} \\
        \cmidrule(lr){2-3} \cmidrule(lr){4-5} \cmidrule(lr){6-7} \cmidrule(lr){8-9} \cmidrule(lr){10-11} \cmidrule(lr){12-13} \cmidrule(lr){14-15} \cmidrule(lr){16-17}
         & \textbf{BL} & \textbf{F1} & \textbf{BL} & \textbf{F1} & \textbf{BL} & \textbf{F1} & \textbf{BL} & \textbf{F1} & \textbf{BL} & \textbf{F1} & \textbf{BL} & \textbf{F1} & \textbf{BL} & \textbf{F1} & \textbf{BL} & \textbf{F1} \\
        \midrule
        
        \rowcolor{gray!10}Claude-4.6-S & 0.79 & 0.68 & 0.71 & 0.52 & 0.72 & 0.49 & 0.36 & 0.59 & 0.42 & 0.64 & 0.40 & 0.54 & 0.42 & 0.54 & 0.575 & 0.737 \\
        \rowcolor{gray!10}GPT-5.4-chat & 0.86 & \textbf{0.78} & \textbf{0.77} & \textbf{0.65} & \textbf{0.78} & \textbf{0.62} & 0.38 & \textbf{0.64} & \textbf{0.43} & 0.67 & 0.41 & \textbf{0.60} & \textbf{0.43} & \textbf{0.63} & 0.589 & 0.740 \\
        \rowcolor{gray!10}Gemini-3.0-f & \textbf{0.87} & \textbf{0.79} & 0.75 & 0.60 & \textbf{0.78} & 0.60 & \textbf{0.40} & \textbf{0.64} & 0.42 & 0.67 & 0.42 & 0.59 & 0.41 & 0.57 & 0.577 & 0.739 \\
        \rowcolor{gray!10}Grok-2-v & 0.70 & 0.66 & 0.55 & 0.42 & 0.48 & 0.34 & 0.23 & 0.48 & 0.28 & 0.51 & 0.35 & 0.53 & 0.30 & 0.48 & 0.562 & 0.720 \\
        \rowcolor{gray!10}Qwen-3.5\textsuperscript{\dag} & \textbf{0.87} & 0.72 & 0.75 & 0.53 & 0.77 & 0.52 & 0.38 & 0.62 & 0.42 & \textbf{0.68} & \textbf{0.46} & \textbf{0.60} & 0.42 & 0.61 & 0.583 & 0.741 \\
        \rowcolor{gray!10}Llama-3.2\textsuperscript{\dag} & 0.85 & 0.81 & 0.78 & 0.73 & 0.74 & 0.69 & \textbf{0.66} & 0.66 & \textbf{0.70} & \textbf{0.70} & \textbf{0.64} & 0.64 & \textbf{0.61} & 0.61 & 0.588 & 0.745 \\
        \rowcolor{gray!10}Gemma-3\textsuperscript{\dag} & 0.72 & 0.63 & 0.60 & 0.43 & 0.58 & 0.41 & 0.52 & 0.52 & 0.58 & 0.58 & 0.50 & 0.44 & 0.49 & 0.47 & 0.570 & 0.725 \\
        \midrule
        \textbf{Non-Reas. Avg} & 0.81 & 0.73 & 0.70 & 0.55 & 0.69 & 0.52 & 0.42 & 0.59 & 0.45 & 0.64 & 0.45 & 0.57 & 0.44 & 0.57 & 0.578 & 0.735 \\
        \midrule
        \rowcolor{cyan!10}GPT-5.4-pro & \textbf{0.92} & \textbf{0.92} & \textbf{0.86} & 0.84 & \textbf{0.84} & 0.79 & \textbf{0.43} & \textbf{0.68} & \textbf{0.46} & \textbf{0.70} & \textbf{0.51} & \textbf{0.70} & \textbf{0.51} & \textbf{0.71} & \textbf{0.593} & \textbf{0.749} \\
        \rowcolor{cyan!10}Gemini-3.0-p & 0.87 & 0.88 & 0.80 & \textbf{0.89} & 0.80 & \textbf{0.86} & 0.39 & \textbf{0.68} & 0.41 & \textbf{0.70} & 0.43 & \textbf{0.70} & 0.43 & 0.66 & 0.586 & 0.747 \\
        \rowcolor{cyan!10}GLM-4.5\textsuperscript{\dag} & 0.84 & 0.83 & 0.78 & 0.80 & 0.77 & 0.76 & 0.41 & 0.66 & 0.44 & 0.68 & 0.46 & 0.66 & 0.45 & 0.66 & 0.591 & 0.747 \\
        \midrule
        \textbf{Reas. Avg} & 0.88 & 0.88 & 0.81 & 0.84 & 0.80 & 0.80 & 0.41 & 0.67 & 0.44 & 0.69 & 0.47 & 0.69 & 0.46 & 0.68 & 0.590 & 0.748 \\
        \bottomrule
    \end{tabular}
\end{table*}

\section{Overall Performance on \ERData}
\label{experiments}

The analyses in Section~\ref{sec:error} identify recurring failure patterns in ERD understanding, including spatial proximity bias, reliance on linguistic cues, and degradation under increasing diagram complexity. We now quantify how broadly these patterns manifest across models and datasets. Table~\ref{tab:final_scores} reports Macro-F1 and BLEU scores across all settings, including the 179-diagram Schema tier of real-world production database schemas, showing consistent performance degradation across difficulty tiers and diagram sources.

We evaluate ten state-of-the-art VLMs under identical zero-temperature settings. Following prior work, we distinguish between \textit{reasoning} models that employ explicit reasoning or reinforcement-based inference mechanisms, and \textit{non-reasoning} models based on standard single-pass decoding. Evaluated systems include commercial models such as ChatGPT-5.4~\cite{ChatGPT}, Gemini-3.0~\cite{Gemini}, Claude-4.6-Sonnet~\cite{Claude}, and Grok-2-Vision~\cite{Grok3}, alongside open-source models including Llama-3.2-90B Vision, Qwen-3.5-35B~\cite{Qwen}, Gemma-3-27B, and GLM-4.5.

\textbf{Observation 1: VLMs Exhibit Partial but Inconsistent ERD Comprehension.} 
Models achieve strong performance on semantic content but weaker consistency on structured outputs. This gap is reflected in higher BLEU scores relative to Macro-F1 (Table~\ref{tab:final_scores}), particularly on curated ERDs. 
On the synthetic dataset, models achieve BLEU scores between 0.69 and 0.81 and Macro-F1 between 0.52 and 0.73. On curated ERDs, BLEU drops to 0.42–0.64, while Macro-F1 remains relatively stable (0.57–0.64). This asymmetry suggests that models extract textual content effectively but struggle with structural consistency. Slightly improved performance on curated ERDs in Silberschatz notation indicates that more regular layouts reduce structural errors. On the Schema tier, BLEU (0.56–0.59) and Macro-F1 (0.72–0.75) track each other closely (Pearson $r=+0.98$), tighter than on any curated or synthetic tier -- consistent with Schema's flat, low-vocabulary JSON structure leaving less room for the semantic/structural split seen elsewhere.
Overall, these results highlight a gap between semantic extraction and structural understanding.

\textbf{Observation 2: VLMs Struggle with EER Elements.} 
Performance varies significantly across ERD components, with strong results on basic elements and substantial degradation on structurally complex ones. Table~\ref{tab:special_f1_scores} reports average F1 scores across EER elements, revealing large disparities. Entities and attributes are recognized reliably, with F1 scores around 0.90 on curated ERDs and 0.95 on synthetic ERDs. In contrast, performance drops for relationships and connectivity by approximately 15\% on curated ERDs and 10\% on synthetic ERDs. More complex elements exhibit sharper declines: multivalued attributes remain below 0.20 F1 for non-reasoning models, N-ary relationships fall below 0.1, and weak entities and relationships show consistently low recognition. Reasoning models improve performance but still struggle on these elements. These results indicate that VLMs perform well on label-driven components but struggle to recover visually encoded structural semantics. Language priors often compensate when visual recognition fails, producing systematic errors on elements such as N-ary relationships, IS-A hierarchies, and derived attributes. This pattern motivates the controlled analyses in Section~\ref{sec:primary}.

The Schema tier offers a useful contrast: despite averaging 46.4 elements per diagram -- far larger than any curated tier -- its Macro-F1 (0.72–0.75) exceeds Web/Instructional performance. Schema diagrams contain zero weak entities, weak relationships, $n$-ary relationships, IS-A hierarchies, or derived/composite/multivalued attributes, since these constructs are not recoverable from relational DDL. With the EER constructs that drive the sharpest degradations above absent by construction, models are scored only on entities, attributes, primary keys, and binary relationships. Size still matters \emph{within} Schema, however: Pearson correlation between element count and Macro-F1 is $-0.66$, and GPT-5.4-pro drops from 0.911 on the 10-element \texttt{movielens} schema to 0.226 on the 579-element \texttt{adventureworks} schema.

\textbf{Observation 3: ERD Size Matters.}
Model performance degrades as diagram complexity increases, with both evaluation metrics showing higher accuracy on simpler ERDs. Non-reasoning models experience a pronounced drop from easy to hard diagrams, while reasoning-capable models are more robust, underscoring their advantage in handling complex structural patterns.
On curated ERDs from \ERData, we group diagrams into five buckets by element count. Across both Chen and Silberschatz notations, performance decreases monotonically with complexity. Macro-F1 drops from $0.80 \rightarrow 0.67$ for Chen ERDs and from $0.82 \rightarrow 0.72$ for Silberschatz ERDs, corresponding to relative declines of $16.8\%$ and $13.9\%$, respectively. The Schema tier shows the same monotonic pattern at larger absolute scale: GPT-5.4-pro's Macro-F1 falls from 0.918 ($\leq$30 elements) to 0.742 (31--60) to 0.364 (61--100) to 0.240 (101+ elements). This trend confirms ERD size is a critical factor in model performance, reflecting limitations in scaling to complex visual structures, independent of whether EER constructs exist.

\paragraph{GED Analysis.} Overall, GED trends are broadly consistent with Fractional F1: reasoning-oriented models achieve lower edit distances, while performance degrades substantially as diagram complexity increases. For example, GPT-5 achieves normalized GED scores of 0.074, 0.119, and 0.157 on Easy, Medium, and Hard ERDs, respectively, while Grok-2-Vision degrades from 0.363 to 0.727 and 0.891 across the same settings. Similarly, IS-A-only diagrams remain challenging despite their simplified structure, with several models exceeding normalized GED values above 1.0. On the Schema tier, GED corroborates F1 even more tightly (Pearson $r=-0.98$ between Macro-F1 and normalized GED) and correlates with element count at $r=+0.63$, with GPT-5.4-pro's mean normalized GED rising from 0.173 ($\leq$30 elements) to 0.862 (101+ elements). We found that GED largely corroborates the trends already captured by Fractional F1 while providing less interpretable component-level insights due to its sensitivity to local structural mismatches in dense ERDs.

\begin{table}
    \centering
    \small 
    \caption{Category-specific F1 Scores for EERs}
    \label{tab:special_f1_scores}
    \begin{tabular}{lcccc}
        \toprule
        & \multicolumn{2}{c}{\textbf{Curated}} & \multicolumn{2}{c}{\textbf{Synthetic}} \\
        \cmidrule(lr){2-3} \cmidrule(lr){4-5}
        \textbf{Type} & \textbf{Non Res} & \textbf{Res} & \textbf{Non Res} & \textbf{Res} \\
        \midrule
        Entities          & 0.89 & 0.93 & 0.97 & \textbf{1.00} \\
        RegAttr           & 0.87 & 0.93 & 0.94 & \textbf{0.97} \\
        Relationships     & 0.74 & 0.87 & 0.89 & \textbf{0.99} \\
        \midrule
        Multivalued       & 0.18 & 0.53 & 0.14 & \textbf{0.59} \\
        Composite         & 0.34 & 0.86 & 0.38 & \textbf{0.92} \\
        Derived           & 0.40 & 0.46 & 0.14 & \textbf{0.76} \\
        N-ary             & 0.25 & \textbf{0.45} & 0.07 & 0.31 \\
        IS-A              & 0.09 & 0.28 & 0.74 & \textbf{0.91} \\
        PrimaryKeys       & 0.67 & 0.78 & 0.95 & \textbf{0.99} \\
        Weak-Ents      & 0.28 & 0.38 & 0.29 & \textbf{0.65} \\
        Weak-Rels & 0.17 & 0.44 & 0.22 & \textbf{0.55} \\
        \midrule
        \textbf{Macro F1} & 0.59 & 0.68 & 0.59 & \textbf{0.80} \\
        \bottomrule
    \end{tabular}
\vspace{-10pt}
\end{table}

\textbf{OCR Accuracy Analysis.} All models achieve high label recognition accuracy (avg.\ 95.5\%, max 99.6\%, min $>$87\%). These results suggest that text extraction is not the primary bottleneck for current VLMs on ERDs.

\textbf{Open-source vs.\ Closed-source.} Open-source models consistently underperform closed-source models, with larger gaps on complex ERDs, suggesting differences in training and reasoning capabilities rather than model size. The Schema tier is a partial exception: GLM-4.5 (open-source) is within 0.002 Macro-F1 of GPT-5.4-pro, and Llama-3.2 outperforms several closed-source models, suggesting the open/closed gap narrows when EER constructs are absent and the task reduces to flat schema extraction.

\textbf{Robustness.} Repeated evaluations over different versions of the models show minimal variation (Macro-F1 differences $<$4\%), indicating stable and reproducible results.

Overall, these findings indicate that ERD understanding is constrained by visual--structural reasoning rather than OCR or stochastic variability, motivating the controlled analyses in Section~\ref{sec:error}.

\section{Conclusion and Discussion}
\label{discussion}

We introduced \textbf{ERUnderstand}, the first large-scale benchmark for structured understanding of Entity--Relationship Diagrams, containing 2,960 diagrams from educational, production, and synthetic sources. Our evaluation across ten state-of-the-art Vision--Language Models shows that while basic ERD elements are recovered reliably (F1: 0.89--1.00), performance degrades substantially on complex Extended ER constructs such as weak entities, multivalued attributes, and $n$-ary relationships. Reasoning-augmented models improve overall Macro-F1 by 15--25\%, but still exhibit systematic limitations in visual-structural reasoning.

Our analysis identifies three recurring failure modes: spatial proximity bias, reliance on linguistic priors over diagram structure, and reduced robustness as diagram complexity increases. These findings demonstrate that understanding ERDs requires more than text extraction or object recognition; it requires recovering the underlying symbolic structure encoded by graphical conventions.

ERUnderstand focuses on widely used ER and Extended ER notations, including weak entities, identifying relationships, and inheritance structures. Future extensions could incorporate richer modeling formalisms, including UML class diagrams and emerging property-graph schema languages~\cite{thalheim2013entity,skavantzos2025entity,beeren2023formal,angles2023pg}, to further expand the scope of multimodal schema understanding.

\textbf{Limitations.} ERUnderstand enables AI-assisted schema documentation, education, and accessibility research, but VLM outputs should not replace human verification in critical settings. Our fractional F1 uses fixed partial-credit weights and greedy label alignment for interpretability; we do not compare against a classical OCR/shape-detection parser; and all evaluations use zero-temperature, single-template prompting. Future work can extend ERUnderstand to broader schema languages, specialized parsing baselines, and stronger constrained-decoding protocols.

\bibliographystyle{ACM-Reference-Format}
\bibliography{sample-base}

\appendix
\clearpage

\section{ERUnderstand Benchmark Details}
\label{sec:dataset_details}

This appendix provides comprehensive technical specifications for the \ERData{} benchmark, detailing the ground-truth data generation process and the prompts used for both dataset creation and VLM evaluation. The benchmark's core innovation lies in pairing each ERD with a precise JSON schema that encodes all structural elements—entities, attributes, relationships, and EER constructs—enabling fine-grained, element-level evaluation.

\subsection{JSON Schema Design and Rationale}

Our JSON format evolved through careful consideration of ERD structural properties. An initial approach using relationship names as dictionary keys was abandoned upon recognizing that diagrams commonly contain multiple relationships with identical names—a fundamental ambiguity that would compromise evaluation integrity.

The final standardized format addresses this through a two-component structure:

\begin{enumerate}[itemsep=2pt]
    \item \textbf{Entities Dictionary}: Entity names serve as keys, with each entry containing lists for regular attributes, primary keys, multivalued attributes, derived attributes, composite attributes (with nested sub-components), and a weak entity flag.
    
    \item \textbf{Relationships Array}: Each relationship is represented as an object containing the relationship name, array of participating entities, corresponding cardinality constraints, and weak relationship flag. For IS-A hierarchies, objects include superclass, subclasses array, and relation name.
\end{enumerate}

This structure ensures unambiguous representation of complex ERD constructs including N-ary relationships, hierarchical structures, and all EER extensions.

\begin{tcolorbox}[colback=blue!5!white,colframe=blue!75!black,
                  breakable,
                  title=\textbf{Ground-Truth JSON Schema Template}]
\begin{lstlisting}[language=json,basicstyle=\ttfamily\footnotesize]
{
  "entities": {
    "Person": {
      "attributes": ["id", "name"],
      "primary_keys": ["id"],
      "multivalued": ["email"],
      "weak": false
    },
    "Student": {
      "attributes": ["student_id", "major"],
      "primary_keys": ["student_id"],
      "derived": ["age"],
      "weak": true
    },
    "Teacher": {
      "attributes": ["teacher_id", "department"],
      "primary_keys": ["teacher_id"],
      "composite": {
        "name": ["first_name", "last_name"]
      },
      "weak": true
    }
  },
  "relationships": [
    {
      "entities": ["Student", "Course"],
      "cardinality": ["1", "N"],
      "name": "takes",
      "weak": false
    },
    {
      "entities": ["Student", "Course", "Semester"],
      "cardinality": ["N", "1", "1"],
      "name": "enrolls",
      "weak": true
    },
    {
      "subclasses": ["Student", "Teacher"],
      "rel_name": "IsA",
      "superclass": "Person"
    }
  ]
}
\end{lstlisting}
\end{tcolorbox}

\subsection{Annotation Protocol and Inter-Annotator Agreement}
\label{sec:annotation-protocol}

To ensure the correctness of the curated benchmark, all real-world ER diagrams were manually converted into the benchmark's standardized JSON representation by two annotators with database backgrounds. Annotation proceeded in two stages. First, the full curated corpus (148 diagrams) was split between the two annotators, who independently annotated their assigned subset while explicitly flagging uncertain elements --- ambiguous relationship participation, entity-versus-attribute decisions, primary-key identification, and notation-specific constructs. Second, all flagged diagrams were jointly reviewed and resolved through discussion, yielding the adjudicated ground-truth annotations released with \textsc{ERUnderstand}.

\paragraph{Double-annotation pilot study.}
To quantify annotation consistency independent of adjudication, we conducted a dedicated double-annotation study on a representative subset of 30 diagrams (15 Web-sourced, 15 Educational-sourced, i.e., textbook diagrams). Both annotators independently labeled the full subset \emph{prior to} any discussion or adjudication. During subsequent review, 21 of the 30 diagrams required correction or discussion, primarily due to inconsistent notation conventions across sources and low-quality or partially blurred source images.

Rather than introducing a separate agreement metric, we measured annotation consistency using the same evaluation pipeline applied throughout \textsc{ERUnderstand}: one annotator's labels were treated as the reference and the other's as the prediction, and agreement was computed as structure-aware Macro-F1 after the standard JSON normalization and partial-credit structural evaluation used for model assessment. This directly measures agreement on the benchmark's target representation, rather than surface- or token-level similarity.

The pilot study yielded a mean diagram-level Macro-F1 agreement of \textbf{89.72\%} across the 30 diagrams, with Web and Educational subsets both near 90\%. Table~\ref{tab:iaa} reports the breakdown by source tier.

\begin{table}[t]
\centering
\caption{Inter-annotator agreement on the double-annotation pilot subset, computed as structure-aware Macro-F1 between independent (pre-adjudication) annotations.}
\label{tab:iaa}
\begin{tabular}{lccl}
\toprule
\textbf{Tier} & \textbf{Diagrams} & \textbf{Mean Macro-F1 (\%)} \\
\midrule
Web             & 15 & 90.25 \\
Educational   & 15 & 89.19 \\
\midrule
Pilot (combined) & 30 & \textbf{89.72} \\
\bottomrule
\end{tabular}
\end{table}

\paragraph{Note on the released benchmark.}
The 148 diagrams released with \textsc{ERUnderstand} follow the split-annotation-plus-adjudication protocol described above, not the independent double-annotation protocol used for the pilot study. Because disagreements identified during adjudication are resolved through discussion before release, the published ground truth is trivially self-consistent by construction; this is not an agreement measurement and should not be interpreted as one. The 89.72\% Macro-F1 from the independent pilot study is therefore the meaningful reliability statistic for this annotation process, and we report it as such.

\subsection{Synthetic Dataset Generation}

The synthetic portion \ERData{} (2,041 of 2,960 diagrams) was generated through a controlled process using constrained LLM generation. This approach ensures consistent quality, comprehensive coverage of EER elements, and semantic validity across diverse domains.

\subsubsection{Prompt Design for Controlled Generation}

The generation process employed carefully designed prompts that enforce strict structural constraints while mandating inclusion of complex EER elements. This dual requirement—limiting basic complexity while requiring advanced features—prevents the LLM from generating overly simplistic schemas and guarantees coverage of visual elements necessary for comprehensive VLM testing.

\begin{tcolorbox}[colback=blue!5!white,colframe=blue!75!black,
                  breakable,
                  title=\textbf{Prompt for Medium Complexity Diagrams}]
\small
\textbf{Instruction:} Give me a json file containing 2 arrays of entities and relationships of an ER diagram for the following context: ``[DOMAIN\_CONTEXT]''. The diagram should include maximum 5 entities, 5 relationships, primary keys, and also may have IsA, ternary relationships, weak entities and relationships, and may have multivalued, derived or composite attributes. The format should be exactly like this: [JSON\_TEMPLATE]
\end{tcolorbox}

\begin{tcolorbox}[colback=blue!5!white,colframe=blue!75!black,
                  breakable,
                  title=\textbf{Prompt for IS-A Hierarchy Diagrams}]
\small
\textbf{Instruction:} Give me a json file containing 2 arrays of entities and relationships of an hierarchical IsA ER diagram for the following context: ``[DOMAIN\_CONTEXT]''. The diagram should include at least 30 entities without any attributes, and just have IsA relationships. The format should be exactly like this: [JSON\_TEMPLATE]
\end{tcolorbox}

The generation process varied constraints by difficulty level:
\begin{itemize}[itemsep=2pt]
    \item \textbf{Easy}: Maximum 4 entities and relationships; no EER elements, IS-A hierarchies, or ternary relationships
    \item \textbf{Medium}: 5+ entities and relationships; optional EER elements including IS-A and multivalued attributes
    \item \textbf{Hard}: 7+ entities and relationships; all EER elements required
    \item \textbf{High}: 20+ entities and relationships; comprehensive EER element coverage
\end{itemize}

The mandatory JSON template enforces canonical output format, essential for consistent automated evaluation. Each generated JSON was manually validated for correctness and semantic coherence before rendering.

\begin{figure*}[t]
    \centering
    \includegraphics[width=0.85\linewidth]{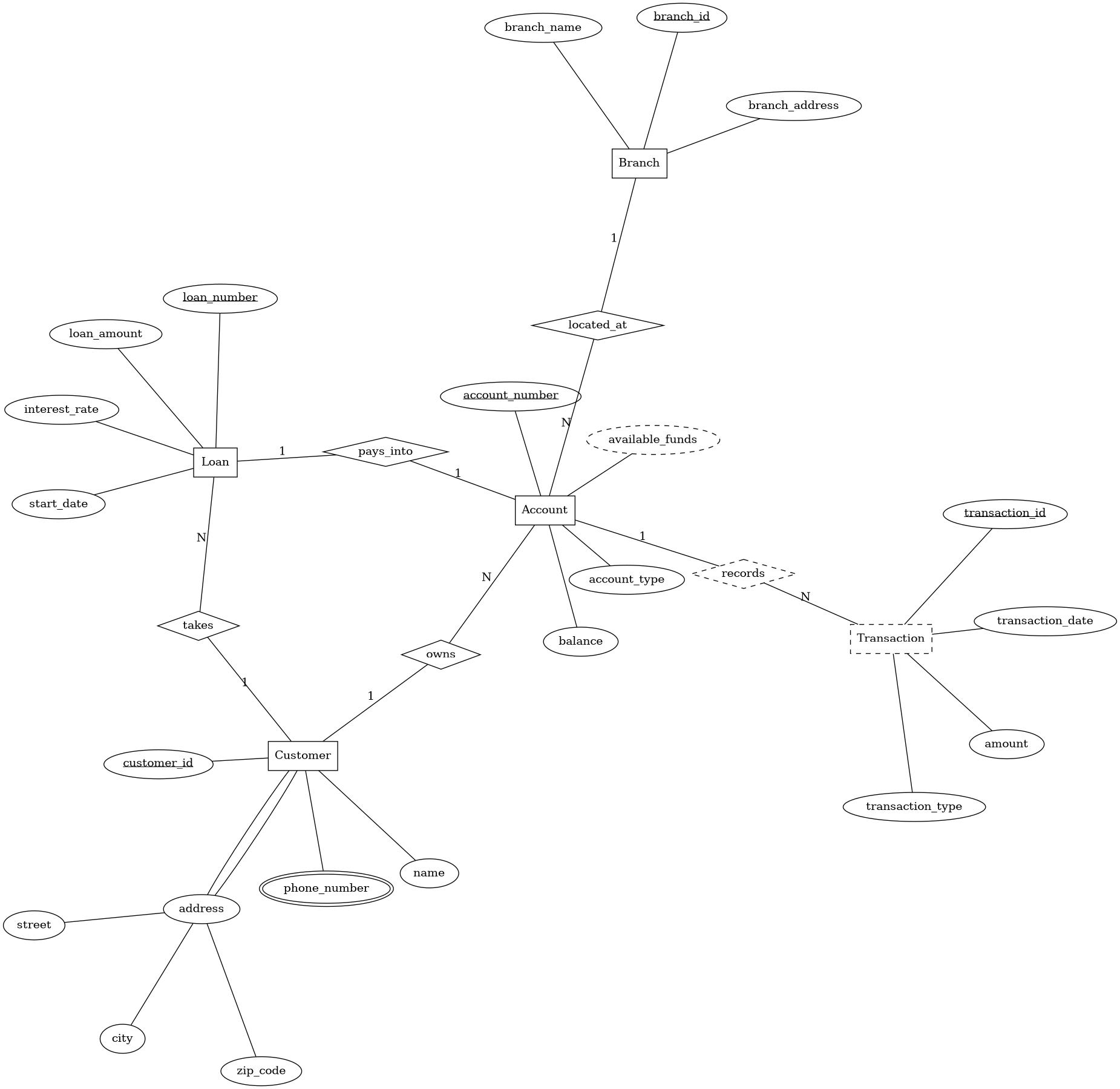}
    \caption{Visual representation of the \texttt{Online Banking} ERD from the synthetic dataset. This diagram illustrates key EER elements: the weak entity \texttt{Transaction} (double rectangle), multivalued attribute \texttt{phone\_number} (double ellipse), derived attribute \texttt{available\_funds} (dashed ellipse), and composite attribute \texttt{address} (hierarchically connected components).}
    \label{fig:erd_example}
\end{figure*}

\section{Evaluation Methodology}
\label{sec:experimental_details}

\subsection{VLM Prompting Strategy}
To ensure fair and consistent evaluation across all tested models, we employed a standardized two-part prompting strategy. This approach minimizes variation due to prompt engineering while testing the VLM's visual comprehension in isolation from its general conversational capabilities.

\begin{tcolorbox}[colback=orange!5!white,colframe=orange!75!black,
                  breakable,
                  title=\textbf{System Prompt}]
\small
\texttt{You are a precise parser. Read the ER diagram image and output ONLY a JSON object that strictly follows the provided format (keys, nesting, and booleans). Do not add explanations, comments, or extra text.}
\end{tcolorbox}

\begin{tcolorbox}[colback=orange!5!white,colframe=orange!75!black,
                  breakable,
                  title=\textbf{User Instruction}]
\small
\texttt{Provide a complete JSON file for this diagram using the schema below: [JSON\_FORMAT\_TEMPLATE]. Do not add any additional text or comments.}
\end{tcolorbox}

The system prompt establishes a constrained output mode, while the user prompt reiterates strict schema adherence requirements. Any deviation in output structure (e.g., added commentary, formatting errors) is counted as a syntax failure, strongly emphasizing the necessity for precise, machine-readable JSON generation based purely on visual content.

\subsection{Text Normalization and Semantic Equivalence}

Our evaluation methodology incorporates several normalization techniques to ensure fair element-level comparison while accounting for minor labeling variations:

\paragraph{Text Normalization}
\begin{itemize}[itemsep=2pt]
    \item \textbf{Punctuation removal}: Characters (\texttt{,}, \texttt{.}, \texttt{-}, \texttt{\_}) stripped from element names
    \item \textbf{Case-insensitivity}: All comparisons performed case-insensitively
    \item \textbf{Whitespace normalization}: Multiple spaces collapsed to single spaces
\end{itemize}

\paragraph{Semantic Equivalence Mapping}
\begin{itemize}[itemsep=2pt]
    \item \textbf{Cardinality normalization}: \texttt{M}, \texttt{N}, \texttt{many}, \texttt{*} treated as equivalent
    \item \textbf{Singular/plural handling}: manually reviewed when normalization leaves semantically distinct forms
    \item \textbf{Orthographic variants}: normalized only for punctuation, case, and spacing; no tunable fuzzy threshold is used in the final scorer
\end{itemize}

Despite these normalization efforts, we observed two categories of persistent semantic errors requiring manual review: (1) inconsistent singular/plural entity naming where semantic meaning differs, and (2) orthographic variations that change semantic meaning (e.g., ``student\_id'' vs. ``students\_identifier''). To maintain high structural standards, any model output that failed strict JSON schema validation (e.g., syntax errors or unclosed brackets) was treated as a null set, resulting in an F1 score of 0 for that instance. Conditional valid-JSON coverage is tracked separately from structural F1 so formatting robustness is not confused with diagram comprehension.

\paragraph{Valid-JSON coverage audit.}
For the paper model set, GPT-5.4-pro, Gemini-3.0-f, and Grok-2-v produced valid JSON for all Internet and Books diagrams. GPT-5.4-chat, Llama-3.2, Gemma-3, and GLM-4.5 had partial archived coverage (46/100 Internet and 21/48 Books), while the remaining missing folders are treated as unavailable rather than structural failures. This audit separates output-format robustness from visual-schema understanding.

\paragraph{Worked $n$-ary relationship example.}
Suppose a diagram has 10 ground-truth relationships and a prediction contains one missed relationship, one extra relationship, one misconnected ternary relationship, and one wrong cardinality. Using Eq.~(1)--(3), $TP_{\text{rel}}=10-1-1-1=7$, $FP_{\text{rel}}=1+(1+1+1)/2=2.5$, and $FN_{\text{rel}}=1+1/2=1.5$, giving $\text{F1}_{\text{rel}}=14/(14+2.5+1.5)=0.778$. Thus a partially correct higher-order relation is penalized, but not counted as a complete failure when its label or participants partially match.

\subsection{Aggregation Strategy for F1-Score}

After computing category-level fractional F1-scores for entities, relationships, attributes, weak entities, weak relationships, and primary keys, we needed an aggregation strategy to produce an overall performance metric. We explored several approaches:

\paragraph{Weighted F1-Score}
We initially experimented with weighting scores according to the frequency of each element type in the diagrams. Although this approach reflected model performance on frequent elements such as entities and attributes, it introduced a ``saturation effect:'' most models achieved very high scores (>0.90) in these categories, obscuring meaningful performance comparisons between models.

\paragraph{Simple and Harmonic Averaging}
We also tested simple averaging and harmonic aggregation. Simple averaging showed promise but still gave disproportionate weight to categories with highly variable performance. Harmonic aggregation, which emphasizes the minimum score, gave too much importance to rare elements such as weak relationships and N-ary relationships. Because these constructs appear infrequently in the dataset, small absolute errors resulted in very low category scores that dominated the aggregate, producing overly pessimistic overall results that did not reflect general model capabilities.

\paragraph{Macro-F1 (Final Choice)}
We ultimately adopted Macro-F1, which computes the unweighted mean of category-level F1-scores. This approach treats all ERD element types as equally important for comprehensive diagram understanding, regardless of their frequency in the dataset. While this means that rare but structurally significant elements (weak entities, N-ary relationships) receive equal weight to common elements (regular entities, attributes), we believe this accurately reflects the goal of complete ERD comprehension. A model that fails systematically on weak entities or N-ary relationships has a fundamental gap in understanding ERD notation, even if these elements appear infrequently.

Macro-F1 provides balanced assessment across all diagram components while avoiding the pitfalls of frequency-based weighting. This choice aligns with our goal of evaluating genuine structural comprehension rather than optimizing for common cases.

\paragraph{Sensitivity to matching and partial-credit choices.}
We additionally re-scored GPT-5 outputs while varying the misconnection partial-credit weight $f$ from 0 to 1. Internet Macro-F1 moved from 0.802 to 0.883 and Books from 0.806 to 0.883, with the default $f=0.5$ yielding 0.835 and 0.834 respectively. This confirms that the main conclusions are not an artifact of a single partial-credit setting; lower weights are stricter, while higher weights forgive misconnections more aggressively.

\paragraph{Additional Graph Edit Distance Results.}
Table~\ref{tab:ged-results} reports normalized Graph Edit Distance (GED) scores across representative subsets of ERUnderstand. Overall, GED trends are broadly consistent with Fractional F1, with reasoning-oriented models generally achieving lower edit distances and better robustness under increasing structural complexity. As ERDs become larger and more densely connected, GED increases substantially across nearly all models, particularly for diagrams containing IS-A hierarchies and higher-order relationships. While GED provides a topology-aware graph-level evaluation signal, we found that it is highly sensitive to small local mismatches, which can propagate into disproportionately large graph penalties in dense diagrams. Consequently, GED primarily serves as a complementary metric to the finer-grained component-level analysis provided by Fractional F1.

\begin{table*}[t]
\centering
\small
\caption{Normalized Graph Edit Distance (GED) across representative ERUnderstand subsets. Lower is better.}
\label{tab:ged-results}
\begin{tabular}{lccccc}
\hline
\textbf{Model} & \textbf{Easy} & \textbf{Medium} & \textbf{Hard} & \textbf{Extreme} & \textbf{IS-A} \\
\hline
Claude-3.5 & 0.150 & 0.306 & 0.297 & 1.096 & 1.066 \\
Claude-4.6 & 0.121 & 0.197 & 0.205 & 1.264 & 0.564 \\
GPT-4o & 0.122 & 0.216 & 0.219 & 1.209 & 1.047 \\
GPT-5 & \textbf{0.074} & 0.119 & 0.157 & 1.334 & 0.926 \\
GPT-5.4-pro & 0.076 & \textbf{0.114} & \textbf{0.101} & -- & -- \\
GPT-5.4-chat & 0.087 & 0.173 & 0.193 & -- & -- \\
Gemini-3.0-flash & 0.123 & 0.308 & 0.262 & 0.476 & 0.639 \\
Gemini-3-Pro & 0.118 & 0.168 & 0.182 & \textbf{0.454} & 1.126 \\
Gemma-3 & 0.302 & 0.500 & 0.467 & -- & -- \\
Grok-2-Vision & 0.363 & 0.727 & 0.891 & 1.300 & 1.728 \\
Llama-3.2 & 0.139 & 0.210 & 0.220 & -- & -- \\
Qwen-3 & 0.112 & 0.291 & 0.263 & 1.176 & \textbf{0.528} \\
Qwen-3.5 & 0.090 & 0.143 & 0.136 & -- & -- \\
\hline
\end{tabular}
\end{table*}

\section{Related Work on Diagram Understanding}
\label{sec:related_details}

\subsection{Diagrammatic Reasoning Foundations}

The term \textit{diagram} encompasses diverse visual representations from statistical plots to informal sketches to rigorous notations like ERDs or UML. Given this breadth, assertions about ``diagram understanding'' often lack precision. Research in diagrammatic reasoning \cite{jamnik2001, stenning1995} examines how visual structures encode and support inference. This field explores VLMs' capabilities with flowcharts \cite{cheng2002}, geometric reasoning \cite{allwein1996}, and data visualizations \cite{Engelhardt}. 

A persistent limitation emerges across these domains: while VLMs can identify entities within diagrams, their ability to reason about relationships between entities remains constrained \cite{hou2024vision}. Models often rely on background knowledge rather than genuine comprehension of visual language structure, and they exhibit tendencies toward hallucination when interpreting diagrammatic content \cite{chen2024, huang2025, manakul2023}.

These challenges connect to the broader field of visual language theory, which examines how shapes, spatial relations, and connectors convey structured meaning according to learned conventions \cite{shin1994}. These conventions—flowchart symbols, geometric figures, relational graphs—were invented to communicate domain-specific reasoning visually \cite{larkin1987}. Understanding such visual grammars demands that models move beyond text recognition to engage in semantic reasoning grounded in spatial logic and symbolic representation \cite{hammer1995}.

\subsection{Heterogeneous Reasoning Systems}

Database diagrams function as formal languages with well-defined syntax, semantics, and reasoning operations. ERDs exemplify this formalism: shapes and connections follow precise compositional rules that generate infinitely many valid instances \cite{chen2024spatialvlm, hsu2023visual, visualstructures2025, barwise1995heterogeneous}. This formal structure enables systematic syntactic validation and error detection, distinguishing ERDs from informal diagrams or workflows lacking standardized conventions \cite{giunchiglia1992theory, shimojima2015semantic}.

The unambiguous encoding of database schemas in ERDs offers an interpretable yet challenging benchmark for distinguishing genuine visual reasoning from pattern matching. By focusing on ERDs, we examine a rich form of diagrammatic reasoning central to data modeling, revealing current VLM limitations in structured visual language comprehension.

\section{VLM Failure Analysis: Detailed Examples}
\label{sec:failure_examples}

This section provides detailed examples of systematic VLM failures, organized by error category. Each example includes the ground-truth structure, model output, and analysis of the failure mechanism.

\subsection{Example 1: EER Element Misclassification}

This example, based on the \textit{Online Banking} context (Figure~\ref{fig:erd_example}), demonstrates common failures in correctly identifying EER elements despite correct basic structure parsing.

\paragraph{Ground Truth Structure}
\begin{itemize}[itemsep=1pt]
    \item Entity \texttt{Account}: Contains derived attribute \texttt{available\_funds} (dashed ellipse)
    \item Entity \texttt{Customer}: Contains multivalued attribute \texttt{phone\_number} (double ellipse)
    \item Entity \texttt{Transaction}: Weak entity (double rectangle border)
\end{itemize}

\paragraph{Detected Errors (Claude 4.6 Output)}
\begin{itemize}[itemsep=2pt]
    \item \textbf{Derived attribute missing}: \texttt{available\_funds} classified as regular attribute
    \item \textbf{Multivalued attribute missing}: \texttt{phone\_number} classified as regular attribute
    \item \textbf{Weak flag mismatch}: \texttt{Transaction} entity weak flag incorrectly set to \texttt{false}
\end{itemize}

\begin{tcolorbox}[colback=purple!5!white,colframe=purple!70!black,
                  breakable,
                  title=\textbf{Claude 4.6 Output (EER Misclassification)}]
\begin{lstlisting}[language=json,basicstyle=\ttfamily\scriptsize]
"Account": {
  "attributes": ["account_number", "available_funds",
                 "account_type", "balance"],
  "primary_keys": ["account_number"],
  "multivalued": [],
  "weak": false
},
"Transaction": {
  "attributes": ["transaction_id", "transaction_date",
                 "amount", "transaction_type"],
  "primary_keys": ["transaction_id"],
  "multivalued": [],
  "weak": false
},
"Customer": {
  "attributes": ["customer_id", "name", "phone_number"],
  "primary_keys": ["customer_id"],
  "multivalued": [],
  "weak": false,
  "composite": {
    "address": ["street", "city", "zip_code"]
  }
}
\end{lstlisting}
\end{tcolorbox}

\paragraph{Analysis}
The model successfully identified the composite attribute (\texttt{address}) but failed on other EER visual notations. This selective success suggests incomplete training on specific visual conventions—double/dashed lines for special attributes and double borders for weak entities. The model appears to recognize hierarchical nesting (composite attributes) more reliably than line-style variations, indicating a specific gap in visual notation understanding rather than a general structural reasoning failure.

\subsection{Example 2: Spatial Reasoning Failure on Dense Diagrams}

Large-scale ERDs with 20+ entities severely stress VLM spatial reasoning and object detection capabilities. Non-Gemini models frequently exhibited near-complete performance collapse on such diagrams.

\begin{figure}[h]
    \centering
    \includegraphics[width=0.9\linewidth]{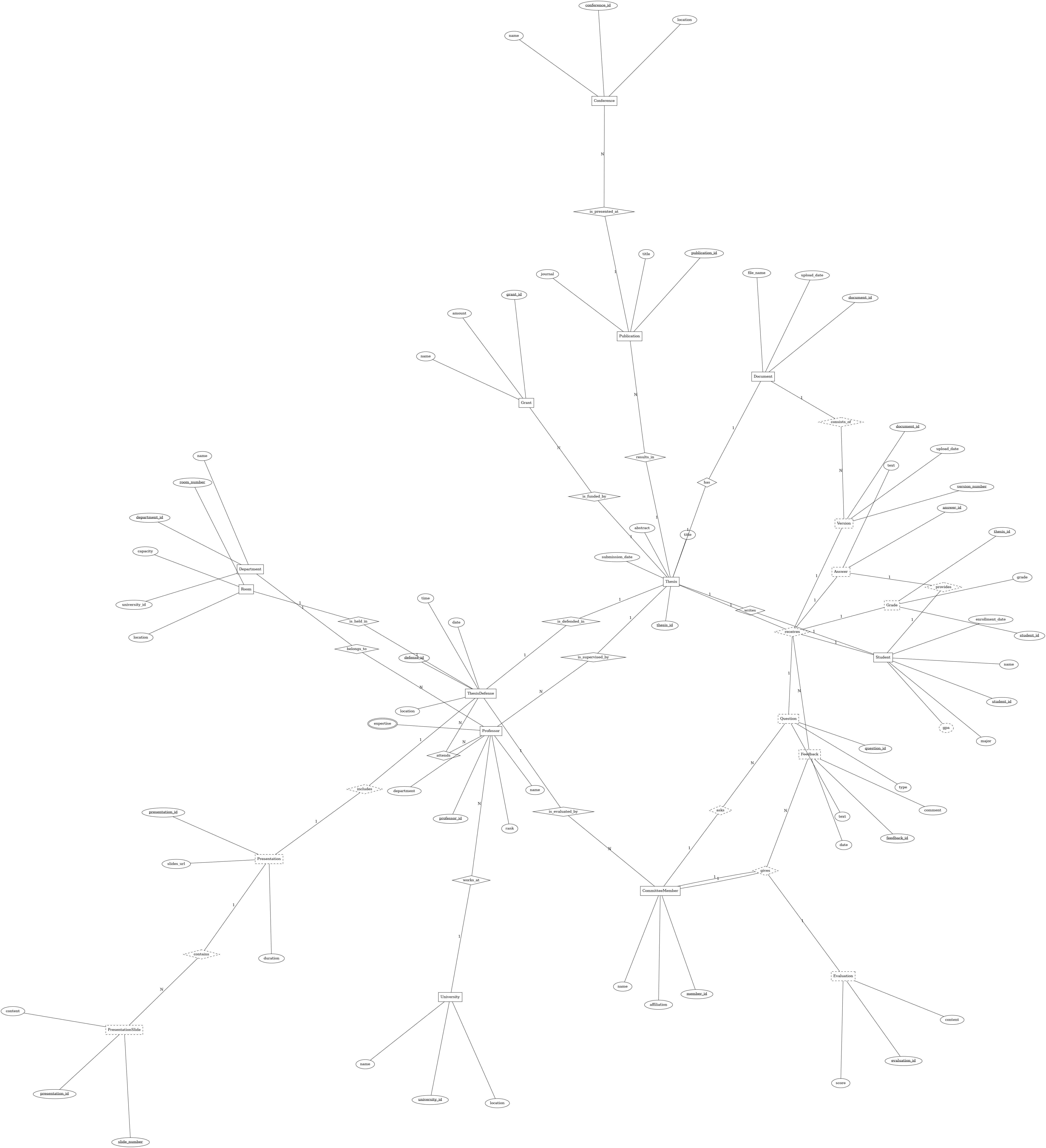}
    \caption{Example of high-density ERD (\texttt{Thesis Defense}) with 22 entities, used to test VLM spatial reasoning under visual complexity. Most models failed to parse beyond 2-3 entities from this diagram.}
    \label{fig:extreme_density}
\end{figure}

\paragraph{Expected Structure}
22 entities including \texttt{Student}, \texttt{Thesis}, \texttt{Committee}, \texttt{Advisor}, \texttt{Department}, etc., connected through 25+ relationships including ternary relationships and weak entity dependencies.

\paragraph{Typical Model Output (GPT-4o)}
\begin{tcolorbox}[colback=purple!5!white,colframe=purple!70!black,
                  breakable,
                  title=\textbf{GPT-4o Output (Dense Diagram)}]
\begin{lstlisting}[language=json,basicstyle=\ttfamily\scriptsize]
{
  "entities": {
    "Student": {
      "attributes": ["student_id", "name"],
      "primary_keys": ["student_id"],
      "weak": false
    },
    "Thesis": {
      "attributes": ["thesis_id", "title"],
      "primary_keys": ["thesis_id"],
      "weak": false
    }
  },
  "relationships": [
    {
      "name": "writes",
      "entities": ["Student", "Thesis"],
      "cardinality": ["1", "1"],
      "weak": false
    }
  ]
}
\end{lstlisting}
\end{tcolorbox}

\paragraph{Analysis}
This partial output captures only 2 of 22 entities and 1 of 25+ relationships, demonstrating near-complete breakdown in visual parsing under high density. The model appears to process only a small, localized region of the diagram—possibly focusing on the most prominent or centered elements—while failing to maintain spatial attention across the full image. Quantitatively, F1-scores dropped below 0.05 across all non-Gemini models for diagrams exceeding 15 entities, confirming that visual complexity represents a critical architectural bottleneck.

\subsection{Example 3: N-ary Relationship Decomposition}

VLMs frequently struggle with ternary and higher-order relationships, often decomposing them into multiple binary relationships or recognizing only a subset of participating entities.

\paragraph{Ground Truth}
Ternary relationship \texttt{schedules} connecting three entities: \texttt{Vet}, \texttt{Pet}, \texttt{Visit} with cardinalities \texttt{[1, N, 1]}.

\paragraph{Model Output (Qwen-3)}
\begin{tcolorbox}[colback=purple!5!white,colframe=purple!70!black,
                  breakable,
                  title=\textbf{Qwen Output (Ternary Collapsed)}]
\begin{lstlisting}[language=json,basicstyle=\ttfamily\scriptsize]
{
  "name": "schedules",
  "entities": ["Vet", "Visit"],
  "cardinality": ["1", "N"],
  "weak": false
}
\end{lstlisting}
\end{tcolorbox}

\paragraph{Analysis}
The model correctly identified the relationship name but collapsed the ternary structure into a binary relationship, completely omitting the \texttt{Pet} entity and its cardinality. This represents a systematic failure mode observed across models (average F1: 0.07 for non-reasoning models on N-ary relationships). The error suggests models struggle with the visual parsing task of tracing multiple connection lines from a single diamond shape, possibly due to attention mechanisms that favor pairwise associations learned from more common binary relationships in training data.

\subsection{Example 4: Composite Attribute Flattening}

Models frequently fail to properly nest composite attribute components, instead flattening all sub-attributes into the main attribute list.

\paragraph{Ground Truth}
Entity \texttt{Speaker} with composite attribute \texttt{name} containing sub-components \texttt{first\_name} and \texttt{last\_name}.

\paragraph{Detected Errors (Gemini-2.0-flash)}
\begin{itemize}[itemsep=2pt]
    \item \textbf{Composite structure missing}: \texttt{name} composite attribute not recognized
    \item \textbf{Flattened components}: \texttt{first\_name} and \texttt{last\_name} appear as regular attributes
    \item \textbf{Redundant parent}: Parent attribute \texttt{name} also appears in regular attributes list
\end{itemize}

\begin{tcolorbox}[colback=purple!5!white,colframe=purple!70!black,
                  breakable,
                  title=\textbf{Gemini-2.0-flash Output (Composite Flattened)}]
\begin{lstlisting}[language=json,basicstyle=\ttfamily\scriptsize]
"Speaker": {
  "attributes": ["speaker_id", "name", "first_name",
                 "last_name", "affiliation", "speech_title"],
  "primary_keys": ["speaker_id"],
  "composite": {},
  "weak": false
}
\end{lstlisting}
\end{tcolorbox}

\paragraph{Analysis}
This error pattern reveals difficulty in recognizing hierarchical attribute structures conveyed through visual nesting of ellipses. The model successfully identified all attribute names but failed to reconstruct their hierarchical relationships. This suggests a limitation in processing spatial containment and hierarchical visual layouts—the model performs well on flat, list-like attribute detection but struggles with the additional inference required to recognize parent-child relationships in composite structures. The presence of all three terms (\texttt{name}, \texttt{first\_name}, \texttt{last\_name}) indicates the model detected the relevant text but misinterpreted the structural semantics.

\subsection{Summary of Failure Patterns}

Our analysis across 1,689 diagrams reveals four primary failure categories:

\begin{enumerate}[itemsep=3pt]
    \item \textbf{Visual notation blindness}: Inability to distinguish dashed/double lines for derived/multivalued attributes and weak entities (F1: 0.14-0.28 for non-reasoning models)
    
    \item \textbf{Spatial complexity collapse}: Near-total failure on diagrams with 20+ entities (F1 < 0.05 for non-Gemini models)
    
    \item \textbf{N-ary relationship decomposition}: Systematic collapse of ternary+ relationships into binary pairs (F1: 0.07 for non-reasoning models)
    
    \item \textbf{Hierarchical structure flattening}: Failure to preserve nesting in composite attributes and IS-A hierarchies despite text recognition (F1: 0.09-0.38)
\end{enumerate}

These patterns indicate that current VLMs excel at text recognition and basic shape detection but struggle with visual reasoning tasks requiring integration of multiple visual cues (line styles, spatial relationships, containment hierarchies). The consistent performance gaps across model architectures suggest fundamental limitations in how VLMs process structured visual languages rather than mere training data deficiencies.

\begin{figure*}[t]
    \centering
    \includegraphics[width=0.30\linewidth]{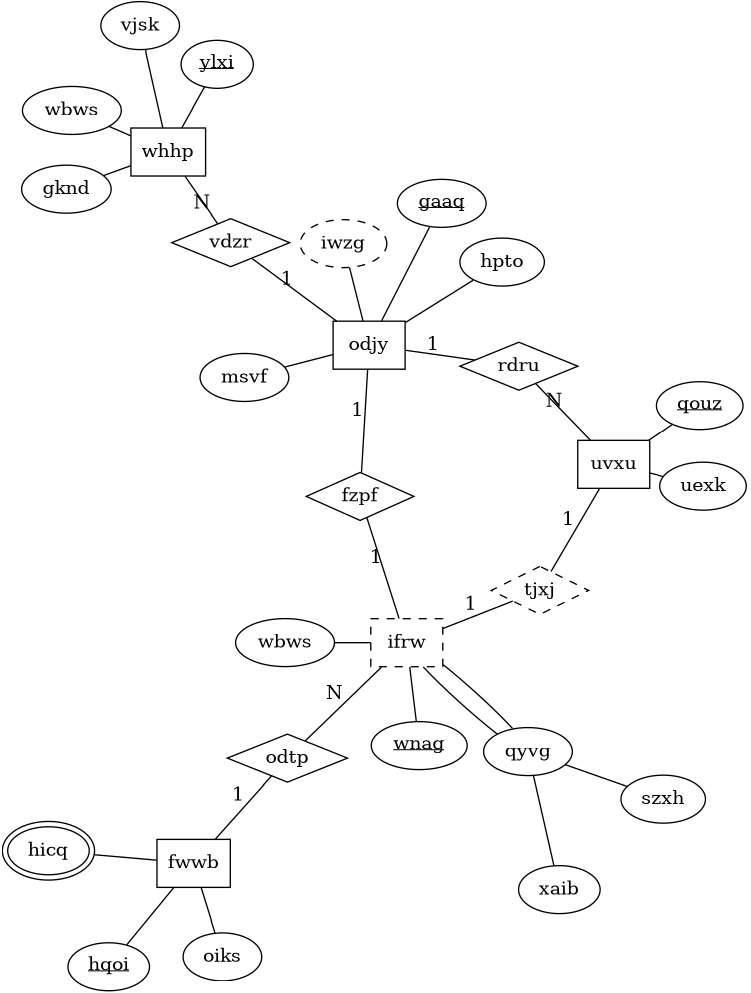}
    \includegraphics[width=0.35\linewidth]{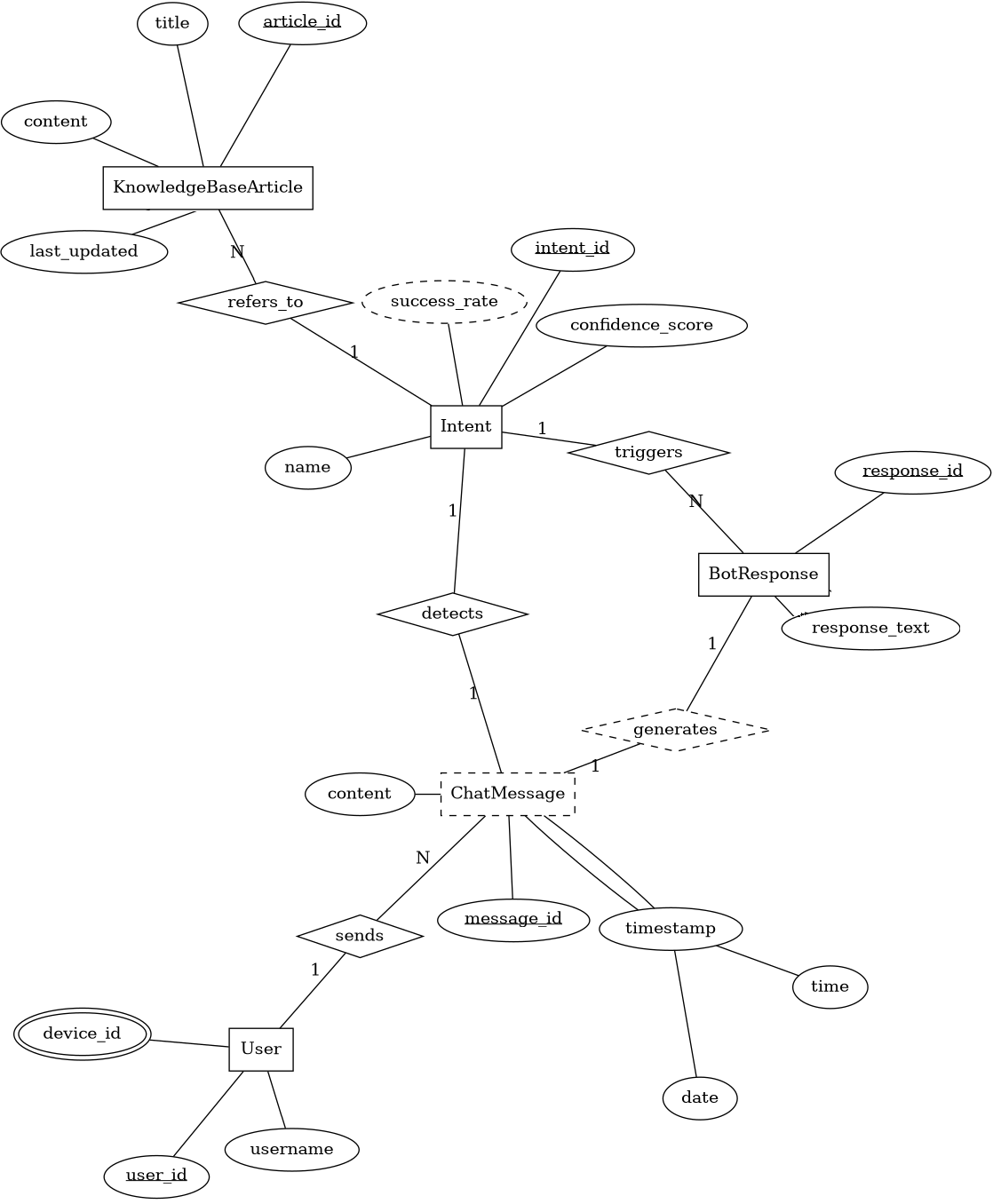}
    \includegraphics[width=0.34\linewidth]{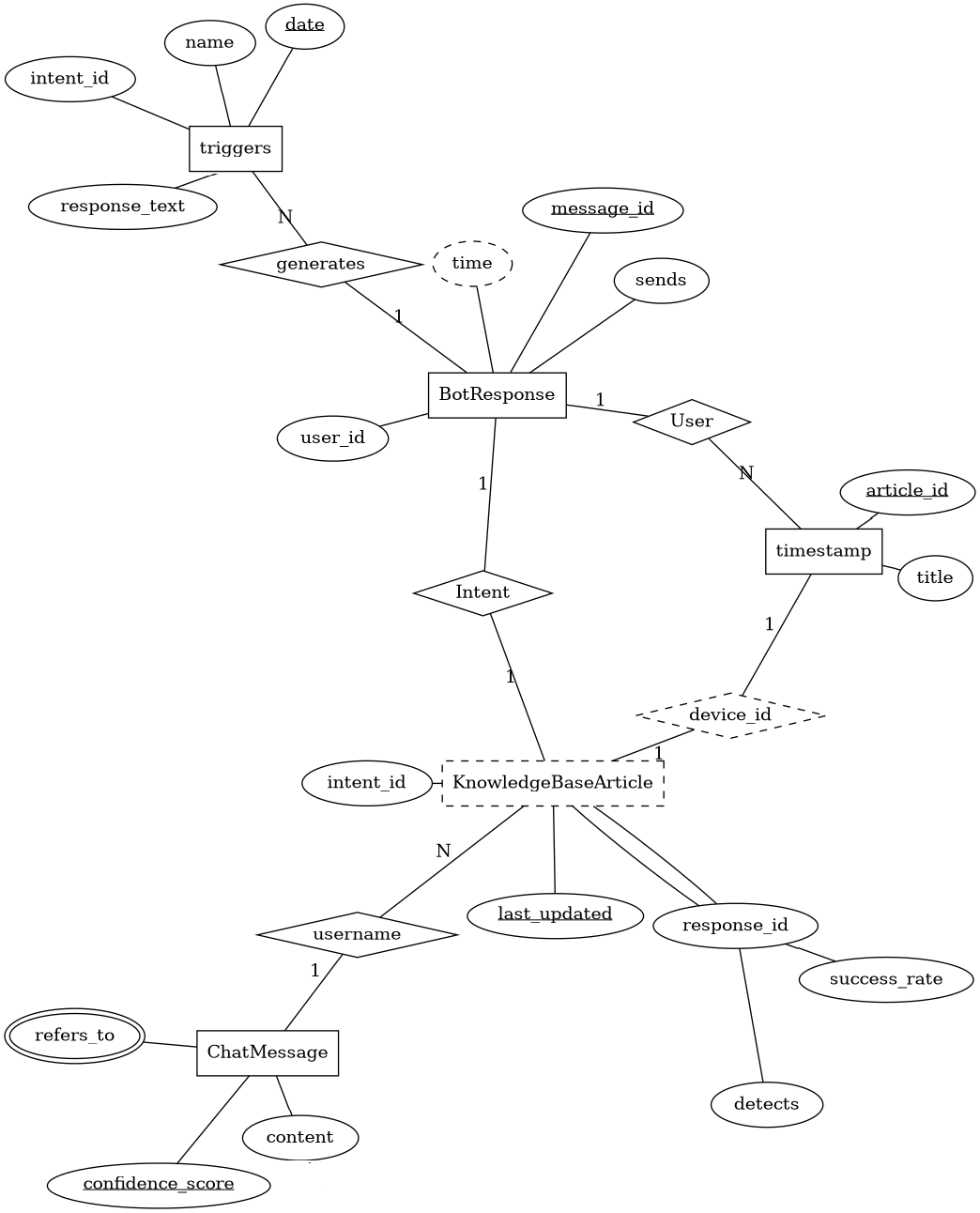}
    \caption{Examples of naming strategies: (left) Context-Free with random 4-letter strings, (middle) Original diagram with semantic labels, (right) Label Permutation with shuffled semantic labels}
    \label{fig:CF_OR_LP}
\end{figure*}

\subsection{Error Mitigation via Interactive Prompting}

\begin{table}[t] \centering \caption{Iterative dialogue interaction error reductions} \begin{tabular}{@{} l @{\;} c @{\;} c @{\;} c @{}} \toprule Model & Initial Errors & Dialog-based & Reduction\% \\ \midrule Grok & 353 & 293 & 17.00 \\ \quad Attr-missed& 81 & 60 & 25.9 \\ \quad PK-extra & 52 & 49 & 5.7 \\ \quad Rel-missed & 14 & 6 & 57.1 \\ \midrule Claude & 203 & 122 & 39.90 \\ \quad Attr-missed& 17 & 12 & 29.4 \\ \quad PK-extra & 38 & 6 & 84.2 \\ \quad Rel-missed & 18 & 12 & 33.3 \\ \midrule Gemini & 228 & 173 & 24.12 \\ \quad Attr-missed& 23 & 14 & 39.1 \\ \quad PK-extra & 61 & 42 & 31.1 \\ \quad Rel-missed & 12 & 7 & 41.7 \\ \bottomrule \end{tabular} \label{tab:dialog} \vspace{-10pt} \end{table}

Having identified systematic biases in spatial reasoning, semantic inference, and visual notation recognition, we investigate whether these errors can be mitigated through iterative interaction.

\textbf{Experimental Setup.}
We select 30 ERDs spanning five difficulty buckets and evaluate three representative models: Grok (lowest-performing), Gemini (second), and Claude (best-performing). Each model is first prompted using a standardized query to extract structured representations, allowing us to identify baseline errors.

\textbf{Interaction Strategy.}
We then engage models in targeted multi-turn dialogue to encourage reassessment. A key challenge is avoiding confirmation bias: directly pointing out errors causes models to overfit to the suggestion. To address this, we adopt an indirect questioning strategy resembling student–teacher interaction. For example, instead of asking “Is CCI a weak relationship?”, we ask “List all weak relationships indicated by double-line diamonds.” This promotes independent reasoning.

\textbf{Results.}
Table~\ref{tab:dialog} summarizes performance changes across interaction rounds. All models improve in recovering missed elements:

\begin{itemize}[leftmargin=*]
    \item \textbf{Grok:} Missed attributes reduced from 0.81 to 0.60 (25.9\%), and missed relationships from 14 to 6 (57.1\%).
    \item \textbf{Claude:} Largest overall improvement (39.9\%), with a substantial reduction in hallucinated PKs (0.38 to 0.06, 84.2\%).
    \item \textbf{Gemini:} Moderate gains, including 41.7\% reduction in missed relationships and 31.1\% reduction in extra PKs.
\end{itemize}

A recurring issue across models is PK hallucination. Upon probing, both Claude and Gemini acknowledge inferring PKs from semantic cues (e.g., “ID”, “email”, “username”) rather than explicit diagram markings.

\textbf{Limitations of Interaction.}
Despite improvements, several error types persist:
\begin{itemize}[leftmargin=*]
    \item \textbf{Structural errors:} Incorrect connections between entities and relationships remain difficult to fix.
    \item \textbf{Notation errors:} Gemini consistently struggles with multivalued attributes and double-line encodings.
    \item \textbf{Instability:} In some cases, additional interaction introduces new errors or degrades performance.
\end{itemize}

\textbf{Discussion.}
These findings reveal that interactive prompting has a selective effect: it is effective for recovering missed elements but less effective for correcting deeper structural or visual reasoning errors. This suggests that interaction primarily enhances attention rather than underlying understanding.

\textbf{Conclusion.}
While iterative prompting improves performance, particularly for omission errors, it cannot fully address limitations in spatial reasoning and visual structure parsing. This highlights the need for architectural improvements beyond prompt engineering.

\subsection{Performance on IS-A Hierarchy Diagrams}
\label{sec:IS-A-appendix}

Observation 2 reveals a large gap in IS-A performance between curated and synthetic ERDs, suggesting reliance on semantic shortcuts. To isolate visual reasoning, we construct 32 IS-A-only diagrams containing only superclass--subclass relationships.

As shown in Table~\ref{tab:IS-A}, performance drops sharply across models: GPT-4o (0.963$\rightarrow$0.118), Gemini-2.5 (0.941$\rightarrow$0.073), and Grok-2-vision (0.519$\rightarrow$0.043). Claude-4.6 is more robust (0.889$\rightarrow$0.415), while Qwen-3 performs best overall (0.919$\rightarrow$0.659).

Despite this, entity recognition remains high (e.g., GPT-5: 0.992, Claude-4.6: 0.994), indicating that failures stem from incorrect hierarchy assignment rather than entity detection. A common error involves reversing superclass and subclass roles.

These findings confirm that IS-A understanding in full ERDs is often supported by semantic context, but degrades significantly when models must rely solely on visual connectivity.

\begin{table}[t]
\centering
\caption{Performance on IS-A-only diagrams compared to medium-complexity diagrams.}
    \label{tab:IS-A}
    \begin{tabular}{llcc}
        \toprule
        \textbf{Model} & \textbf{Dataset} & \textbf{F1-Entities} & \textbf{F1-IS-A} \\
        \midrule
        \rowcolor{cyan!10}
        Grok-2-vision & Medium & 0.874 & 0.519 \\
        \rowcolor{cyan!10}
        Grok-2-vision & IS-A-only & 0.624 & 0.043 \\
        \rowcolor{gray!10}
        Qwen-3 & Medium & 0.977 & 0.919 \\
        \rowcolor{gray!10}
        Qwen-3 & IS-A-only & 0.956 & 0.659 \\
        \rowcolor{cyan!10}
        Gemini-3.0-flash & Medium & 0.981 & 0.904 \\
        \rowcolor{cyan!10}
        Gemini-3.0-flash & IS-A-only & 0.956 & 0.468 \\
        \rowcolor{gray!10}
        GPT-5.4 & Medium & 0.995 & 0.963 \\
        \rowcolor{gray!10}
        GPT-5.4 & IS-A-only & 0.678 & 0.118 \\
        \rowcolor{cyan!10}
        Claude-3.5 & Medium & 0.991 & 0.570 \\
        \rowcolor{cyan!10}
        Claude-3.5 & IS-A-only & 0.412 & 0.187 \\
        \midrule
        \rowcolor{gray!10}
        Gemini-2.5 & Medium & 0.998 & 0.941 \\
        \rowcolor{gray!10}
        Gemini-2.5 & IS-A-only & 0.890 & 0.073 \\
        \rowcolor{cyan!10}
        GPT-5.4-pro & Medium & 0.999 & 0.941 \\
        \rowcolor{cyan!10}
        GPT-5.4-pro & IS-A-only & 0.992 & 0.104 \\
        \rowcolor{gray!10}
        Claude-4.6 & Medium & 0.998 & 0.889 \\
        \rowcolor{gray!10}
        Claude-4.6 & IS-A-only & 0.994 & 0.415 \\
        \bottomrule
    \end{tabular}
\end{table}

\subsection{PKs: Visual vs.\ Naming Conventions Bias}
\label{sec:primary-appendix}

Although PK identification appears strong overall, we observe a significant performance gap between curated and synthetic datasets, suggesting reliance on naming conventions. In synthetic data, PKs consistently follow the \texttt{\_id} pattern, raising the question of whether models use visual or linguistic cues.

We design controlled experiments to isolate this effect: modifying PK names (Changed-PK), removing underline notation (Removed-PK), introducing misleading names (LP), and randomizing attribute names (CF).

Table~\ref{tab:pks} shows that altering naming conventions significantly degrades performance, while removing visual cues leads to widespread PK hallucination. Misleading naming produces the worst results, confirming strong reliance on linguistic patterns.

Dataset statistics further reveal an imbalance: only 45\% of curated entities explicitly mark PKs, compared to dense PK annotation in synthetic data. This amplifies the naming bias and explains cross-dataset performance gaps.

Overall, results demonstrate that VLMs fail to reliably use visual PK notation and instead depend heavily on naming conventions, limiting their robustness in real-world ERDs.

\begin{table}[t]
\centering
\caption{Average PK identification errors across all VLMs.}
\label{tab:pks}
\begin{tabular}{lcc}
\toprule
\textbf{Study} & \textbf{PrimKey-missed} & \textbf{PrimKey-extra} \\
\midrule
\rowcolor{cyan!10}Curated & 61.83 & 161.52 \\
\rowcolor{cyan!10}Synthetic & 40.75 & 38.04 \\
Changed-PK & 105.50 & 56.25 \\
Removed-PK & 0.00 & 479.62 \\
LP & 979.88 & 799.25 \\
CF & 799.00 & 638.12 \\
\bottomrule
\end{tabular}
\end{table}

\subsection{Diagram Size Limits}
\label{sec:size-appendix}

Observation 3 showed performance degradation as diagram complexity increases. To further probe scalability limits, we construct 19 high-complexity ERDs, each containing at least 20 entities and 20 relationships, along with dense connectivity and full EER features.

\textbf{Results.}
As shown in Table~\ref{tab:extreme_complexity}, most models exhibit near-complete failure. Non-Gemini models achieve Macro-F1 scores between 0.035 and 0.048, with relationship extraction particularly weak (e.g., 0.009 for Qwen-3). Notably, even recent reasoning models such as GPT-5 and Claude-4.6 do not improve over earlier versions.

In contrast, Gemini models demonstrate strong robustness. Gemini-2.0-flash achieves 0.428 Macro-F1, with high entity (0.929) and relationship (0.734) scores, while Gemini-2.5 further improves to 0.695, performing consistently well across all components.

\textbf{Discussion.}
These results suggest a fundamental scalability limitation in most VLMs when processing densely structured diagrams. The consistent failure across architectures indicates a likely bottleneck in spatial reasoning and long-range visual dependency tracking rather than data availability.

Gemini’s superior performance points to potential architectural differences—such as improved attention mechanisms or visual encoding strategies—that better support large, complex visual inputs. Given that real-world ERDs often exhibit similar complexity, these findings highlight a critical limitation for practical deployment.

\begin{table}[t]
\centering
\caption{Performance on high complexity diagrams, showing a superior performance of Gemini models in this task.}
\label{tab:extreme_complexity}
\begin{tabular}{lcccc}
\toprule
\textbf{Model} & \textbf{Macro-F1} & \textbf{Ents} & \textbf{Attrs} & \textbf{Rels} \\
\midrule
\multicolumn{5}{l}{\textit{Non-reasoning}} \\
Grok-2-vision & 0.044 & 0.033 & 0.149 & 0.020 \\
Qwen-3 & 0.040 & 0.037 & 0.146 & 0.009 \\
\textbf{Gemini-3.0-flash} & \textbf{0.428} & \textbf{0.929} & \textbf{0.819} & \textbf{0.734} \\
GPT-5.4 & 0.045 & 0.063 & 0.157 & 0.045 \\
Claude-3.5 & 0.035 & 0.035 & 0.143 & 0.019 \\
\midrule
\multicolumn{5}{l}{\textit{Reasoning}} \\
\textbf{Gemini-3-pro} & \textbf{0.695} & \textbf{0.986} & \textbf{0.819} & \textbf{0.766} \\
GPT-5.4-pro & 0.035 & 0.041 & 0.155 & 0.032 \\
Claude-4.6 & 0.048 & 0.105 & 0.171 & 0.040 \\
\bottomrule
\end{tabular}
\vspace{-10pt}
\end{table}

\subsection{Performance on EER elements}

Category-specific F1 scores reveal how VLMs handle different EER elements and expose their primary weaknesses. These elements are central to database design, and reliably recognizing them is essential. Table~\ref{tab:special_f1_scores} presents average F1 scores across ER elements, showing large disparities that warrant targeted investigation.

\textbf{Entities, Relationships, and Attributes.}
Entities and attributes are fundamental ERD primitives, and models recognize them well, achieving approximately 0.90 F1 on curated ERDs and 0.95 on synthetic ERDs. In contrast, performance drops for relationships and connectivity. On curated ERDs, relationship-related scores decrease by roughly 15\% points relative to entity-recognition, and by about 10\% points on synthetic ERDs. This pattern indicates that models often succeed on label-driven components but struggle with the visual structure that encodes relational semantics.

\textbf{Multivalued Attributes.}
Multivalued attributes represent fields with multiple values per entity instance. In Chen notation they appear as double ellipses, and in Silberschatz-style notation they are shown with brackets. Performance remains below 0.20 F1 for non-reasoning models but improves to around 0.60 for reasoning models.

\textbf{Composite Attributes.}
Composite attributes decompose into sub-attributes, typically represented as hierarchically connected ellipses. Non-reasoning models achieve below 0.40 F1, whereas reasoning models improve substantially to values between 0.85 and 0.90 on curated and synthetic ERDs.

\textbf{Derived Attributes.}
Derived attributes are computed from others and are represented as dotted ellipses in Chen notation and parentheses in Silberschatz-style notation. Non-reasoning models perform poorly on synthetic ERDs, with an F1 of 0.14, showing limited sensitivity to this notation.

\textbf{N-ary Relationships.}
N-ary relationships involve three or more entities and require tracing multiple connections simultaneously. Performance drops sharply, remaining below 0.01 F1 for non-reasoning models and reaching only 0.31 for reasoning models.

\textbf{Is-A Relationships.}
Is-A hierarchies encode inheritance and are drawn as triangles in Chen notation and indicated using directed edges in Silberschatz-style notation. The extreme disparity between curated and synthetic ERDs suggests that models may leverage semantic cues rather than tracing the inheritance structure. This motivates our dedicated hierarchy tracing study in Section~\ref{sec:IS-A}.

\textbf{Primary Keys.}
Primary keys uniquely identify entity instances and are typically marked by underlining. Although detection appears relatively strong overall, a substantial gap exists between curated and synthetic ERDs. Since synthetic primary keys frequently follow the \texttt{\_id} naming convention, we evaluate whether models rely on visual cues or naming patterns in Section~\ref{sec:primary}.

\textbf{Weak Entities and Weak Relationships.}
Weak entities and their identifying relationships rely on subtle boundary cues. Recognition remains low, highlighting limited sensitivity to fine-grained visual distinctions. 

Overall, VLMs consistently underperform on visually subtle or structurally complex notation. Language priors often compensate when visual recognition fails, producing systematic errors on elements such as N-ary relationships, Is-A hierarchies, and derived attributes. These patterns motivate the controlled error analyses presented in Sections~\ref{sec:IS-A} through~\ref{sec:primary}.

\end{document}